\documentclass[10pt,twocolumn,letterpaper]{article}

\usepackage[pagenumbers]{cvpr} 

\usepackage{booktabs}
\usepackage{graphicx,amsmath,bbding,amsbsy,amsfonts,amssymb,caption,subcaption,multirow,multicol,overpic,textpos,makecell,url,wrapfig,booktabs,nicefrac,microtype,xspace,array,algorithm,algpseudocode,enumitem,bm,diagbox}
\usepackage[table]{xcolor}

\def\etal{\emph{et al}\onedot}

\usepackage{pifont}
\newcommand{\cmark}{\ding{51}}%
\newcommand{\xmark}{\ding{55}}%

\newcolumntype{x}[1]{>{\centering\arraybackslash}p{#1pt}}
\newcolumntype{y}[1]{>{\raggedright\arraybackslash}p{#1pt}}

\definecolor{deemph}{gray}{0.6}

\definecolor{baselinecolor}{gray}{.9}
\newcommand{\baseline}[1]{\cellcolor{baselinecolor}{#1}}
\newlength\savewidth\newcommand\shline{\noalign{\global\savewidth\arrayrulewidth
		\global\arrayrulewidth 1pt}\hline\noalign{\global\arrayrulewidth\savewidth}}
\newcommand{\tablestyle}[2]{\setlength{\tabcolsep}{#1}\renewcommand{\arraystretch}{#2}\centering\footnotesize}

\usepackage[pagebackref,breaklinks,colorlinks]{hyperref}

\usepackage[capitalize]{cleveref}
\crefname{section}{Sec.}{Secs.}
\Crefname{section}{Section}{Sections}
\Crefname{table}{Table}{Tables}
\crefname{table}{Tab.}{Tabs.}

\def\cvprPaperID{1888} 
\def\confName{CVPR}
\def\confYear{2023}

\begin{document}

\title{\emph{FastMIM}: Expediting Masked Image Modeling Pre-training for Vision}

\author{Jianyuan Guo$^{1,2}$, Kai Han$^{2}$, Han Wu$^{1}$, Yehui Tang$^{2}$, Yunhe Wang$^{2}$, Chang Xu$^{1}$ \\
	\normalsize$^1$ School of Computer Science, Faculty of Engineering, University of Sydney.
	\normalsize$^2$ Huawei Noah’s Ark Lab.
	\\
	\small\texttt{jianyuanguo12138@gmail.com; \{kai.han, yunhe.wang\}@huawei.com; c.xu@sydney.edu.au}
}

\maketitle

\begin{abstract}
The combination of transformers and masked image modeling (MIM) pre-training framework has shown great potential in various vision tasks. However, the pre-training computational budget is too heavy and withholds the MIM from becoming a practical training paradigm. This paper presents FastMIM, a simple and generic framework for expediting masked image modeling with the following two steps: (i) pre-training vision backbones with low-resolution input images; and (ii) reconstructing Histograms of Oriented Gradients (HOG) feature instead of original RGB values of the input images.
In addition, we propose FastMIM-P to progressively enlarge the input resolution during pre-training stage to further enhance the transfer results of models with high capacity.
We point out that: (i) a wide range of input resolutions in pre-training phase can lead to similar performances in fine-tuning phase and downstream tasks such as detection and segmentation; (ii) the shallow layers of encoder are more important during pre-training and discarding last several layers can speed up the training stage with no harm to fine-tuning performance; (iii) the decoder should match the size of selected network; and (iv) HOG is more stable than RGB values when resolution transfers;. Equipped with FastMIM, all kinds of vision backbones can be pre-trained in an efficient way. For example, we can achieve 83.8\%/84.1\% top-1 accuracy on ImageNet-1K with ViT-B/Swin-B as backbones. Compared to previous relevant approaches, we can achieve comparable or better top-1 accuracy while accelerate the training procedure by $\sim$5$\times$. Code\footnote{Mindspore~\cite{mindspore} implementation: \href{https://toscode.gitee.com/mindspore/hub/blob/master/mshub_res/assets/noah-cvlab/gpu/1.8/fastmim_v1.0_imagenet2012.md}{https://toscode.gitee.com/mindspore/hub}.} can be found in \href{https://github.com/ggjy/FastMIM.pytorch}{https://github.com/ggjy/FastMIM.pytorch}.
\end{abstract}

\begin{figure}
	\centering         
	\includegraphics[width=\linewidth]{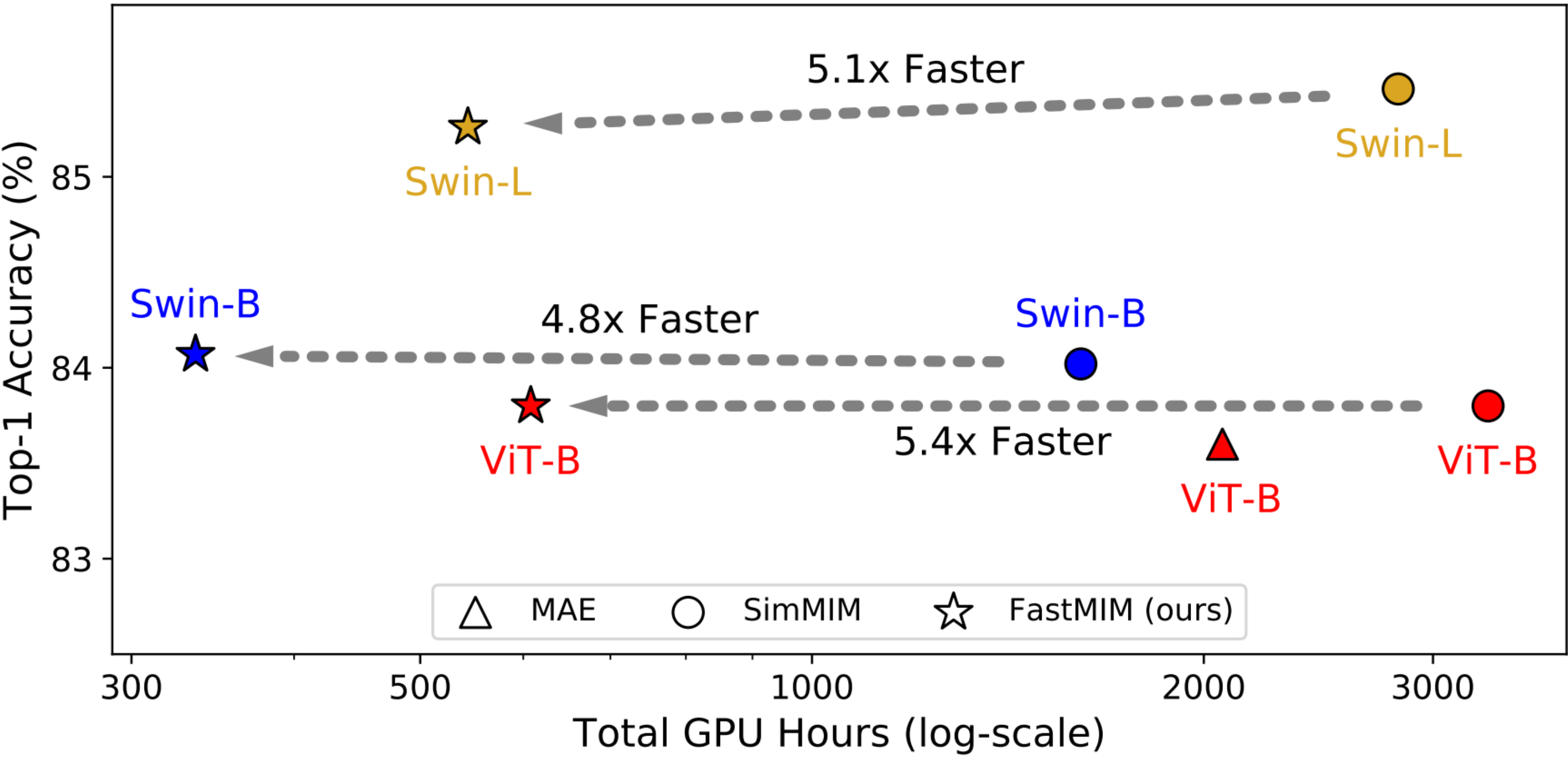}
	\vspace{-0.5cm}
	\caption{\small{Comparisons in terms of total GPU hours (pre-training time) on ImageNet-1K classification task. $\bigtriangleup$ indicates pre-trained by MAE~\cite{mae}, \large $\circ$ \small indicates SimMIM~\cite{simmim}, and \normalsize \text{\ding{73}} \small indicates our \emph{FastMIM}. Encoders \textcolor{red}{ViT-B}/\textcolor{blue}{Swin-B}/\textcolor{brown}{Swin-L} are shown in different colors. \emph{FastMIM} expedites the training stage by $\sim$5$\times$.}}
	\vspace{-0.2cm}
	\label{fig:intro}
\end{figure}

\section{Introduction}
Self-supervised learning aims to learn feature representations from scalable unlabeled data. It has obtained significant achievements in the field of natural language processing (NLP) through masked language modeling (MLM)~\cite{radford2018improving,bert,brown2020language,igpt}, and has also attracted increasing attention in computer vision. Driven by the great success of MLM, masked image modeling (MIM) emerged as a promising self-supervised pre-training paradigm for vision community. Different from previous contrastive learning based approaches~\cite{wu2018unsupervised,simclr,mocov1,dino}, MIM learns representations through a mask-then-predict manner, \eg, predicting the raw pixels~\cite{mae,simmim} or other tokenizations~\cite{beit,ibot,cae} of the randomly masked input images.

\begin{figure*}[t]
	\centering         
	\includegraphics[width=\textwidth]{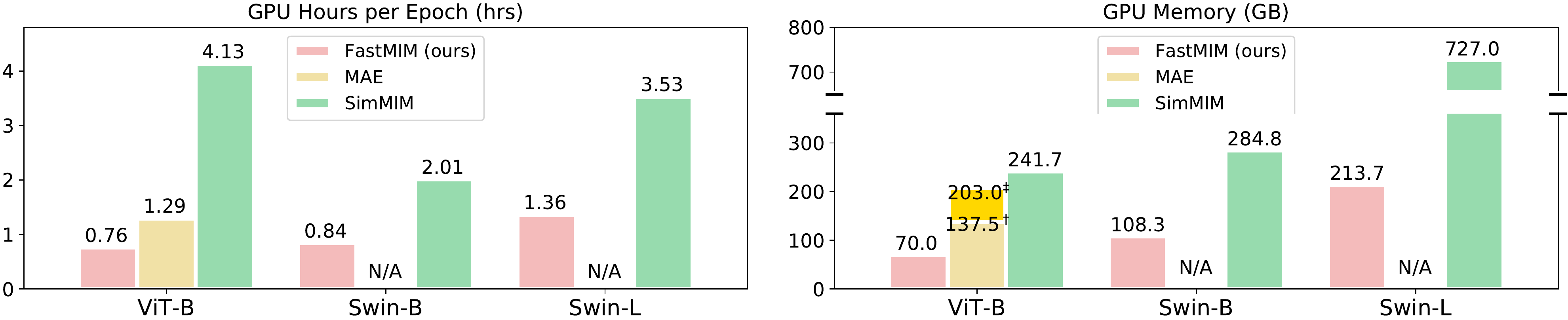}
	\vspace{-0.5cm}
	\caption{\small{Comparison of our \emph{FastMIM}, MAE~\cite{mae} and SimMIM~\cite{simmim} in terms of GPU efficiency. All frameworks use a ViT-B/Swin-B/Swin-L encoder and a batch size of 2048. The experiments are conducted on a single machine with 8 32GB V100 GPUs, CUDA 10.1, and PyTorch 1.8.1. ($^\dagger$: MAE decoder has 1 block (1b512d). $^\ddagger$: MAE decoder has 8 blocks (8b512d). N/A: MAE is not suitable for Swin Transformer.)}}
	\vspace{-0.2cm}
	\label{fig:gpu_efficiency}
\end{figure*}

Although self-supervised learning based approaches have recently shown great promise and achieved state-of-the-art results on various downstream vision tasks, the pre-training stage is extremely \textbf{\emph{slow}} and \textbf{\emph{computationally expensive}}. For example, it takes contrastive learning based SimCLR~\cite{simclr} 15 hours on 128 TPU v3 cores (1920 TPU hours in total) to finish the 1000 epochs training on ResNet-50~\cite{resnet} with a batch size of 4096. Moreover, MIM based BEiT~\cite{beit} takes about five days using 16 32GB V100 GPUs (1920 GPU hours in total, not counting the time for dVAE~\cite{dvae,vqvae} pre-training) to accomplish 800 epochs training on ViT-B~\cite{vit}. To pre-train vision backbones efficiently, He~\etal proposes the masked autoencoder (MAE)~\cite{mae} which discards the masked tokens and only operates on the whole input sequences in the lightweight decoder. Notably, although this asymmetric encoder-decoder design significantly reduces the computational burden, MAE can only support the isotropic transformer architecture~\cite{vit} as the encoder, withholding it from becoming a \textbf{\emph{generic}} MIM framework for various vision backbones~\cite{pvt,swin,cmt,twins}. In contrast to above discarding strategy, SimMIM~\cite{simmim} retains both visible and masked tokens. In this way, SimMIM can be naturally applied to different architectures, \eg, isotropic ViT~\cite{vit} and hierarchical Swin Transformer~\cite{swin}. However, it suffers from heavy memory consumption that even the base size model of Swin Transformer (Swin-B) cannot be trained via SimMIM framework on a single machine with 8 32GB V100 GPUs~\cite{greenmim}.

To reduce the pre-training costs of self-supervised learning and make MIM a \emph{generic} framework for vision community, we devise a simple and straightforward framework (Figure~\ref{fig:fastmim}), viz, \emph{FastMIM}, for faster training speed and easier deployment of AI applications. Inspired by SimMIM~\cite{simmim} that retains all input tokens during pre-training stage, we directly mask on the raw RGB input and keep the illuminated input the same as in the supervised learning producer. This presents a fresh opportunity for \emph{FastMIM} to serve as a \textbf{\emph{generic}} framework because no modification is made to the architecture and the input shape. Yet, standard input images of size 224$\times$224 are inherently used in pre-training stage in common practice~\cite{beit,mae}. For example, given the 16$\times$16 patch size, the encoder of ViT-B~\cite{vit} need to tackle with 196 input patches in SimMIM~\cite{simmim}. To alleviate the memory consumption, we come up with a straightforward approach of \emph{reducing the input resolution}, \eg, from 224$\times$224 to 128$\times$128, and the number of input patches are reduced to 64 accordingly, as shown in Figure~\ref{fig:gpu_efficiency} and~\ref{fig:fastmim}. We further leverage the HOG target~\cite{hog,maskfeat} to compensate for the loss of texture information resulted from the reduction of image resolution. Our main contributions can be summarized as:
\begin{itemize}[itemsep=0.1pt, topsep=0.1pt, leftmargin=*]
	\item We investigate various configurations of MIM framework, and identify the \emph{key} design to expedite the pre-training stage and reduce the memory consumption: directly \textbf{\emph{reducing the input resolution}} for MIM.
	\item We elaborate the characteristic of the \textbf{\emph{HOG feature}}, which is \textbf{\emph{almost invariant to the geometric changes in images}}. Compared to pixel target, reconstructing HOG target can better compensate for the loss of texture information resulted from the reduction of image resolution.
	\item Based on the above observations, we propose the \emph{FastMIM} which can expedite overall pre-training speed by 5$\times$, and reduce the memory consumption simultaneously. Extensive experiments demonstrate the effectiveness and efficiency of our proposed framework in practical AI fields.
\end{itemize}

\noindent Overall, the heavy memory consumption of previous self-supervised learning frameworks erects an unfortunate barrier for more researchers to dive into this filed. We hope our findings and the proposed \emph{FastMIM} can provide avenues and insights for the MIM pre-training for vision community.

\section{TL;DR for Implementation}
As shown in Figure~\ref{fig:fastmim}, it is simple to re-implement our \emph{FastMIM} based on SimMIM~\cite{simmim} and MaskFeat~\cite{maskfeat}. More specifically, we directly execute ``masking procedure" on original input images, and then reduce the input resolution during pre-training stage. We leverage the HOG descriptor as the prediction target. Low input resolution and HOG target are two most crucial factors to expedite MIM procedure while maintaining the high performance. More analyses on pixel and HOG target are shown in Sec.~\ref{sec:abla_on_target} and Sec.~\ref{appendix_sec:abla_hog}.

\section{Revisit Masked Image Modeling}

Our \emph{FastMIM} is a simple and straightforward framework based on masked image modeling~\cite{simmim}, which masks a portion of original images, and predicts the masked regions. We start by revisiting MIM and elaborate our observations.

\subsection{Preliminary}
\vspace{-0.1cm}
\noindent\textbf{Notations.} Following the commonly used configurations~\cite{mae,beit,simmim}, given an input image $\mathbf{X} \in \mathbb{R}^{H\times W\times C}$, where $H$, $W$, and $C$ are the height, width, and number of channels, some of the pixels in $\mathbf{X}$ are randomly masked out by being replaced with a mask token, denoted as [M]. Let $\mathbf{S} \in \{0, 1\}^{H\times W\times C}$ denotes the spatial mask, where $0$ indicates a pixel\footnote{Here we directly mask on the input RGB images, which is different with previous methods~\cite{simmim,beit,greenmim} which mask on the image patches (tokens).} is invisible for the encoder, and $1$ indicates a pixel is visible.

\begin{table*}[t]
	\centering
	{\begin{subfigure}{0.35\linewidth}
			\begin{center}
				\small
				\tablestyle{4.5pt}{1.2}
				\begin{tabular}{cccccc}
					case & mask at & encoder & Top-1 & AP$^{\bf b}$ & AP$^{\bf m}$ \\ 
					\shline
					MAE & patches & w/o [M] & 83.6 & 50.3 & 44.9 \\
					MIM & patches & w/ [M] & 83.7 & 50.5 & 45.1 \\
					\baseline{MIM} & \baseline{image} & \baseline{w/ [M]} & \baseline{83.7} & \baseline{50.4} & \baseline{45.1} \\
				\end{tabular}
				\caption{\textbf{Mask strategy.} The encoder is ViT-B and the setting of decoder is 4b512d.}
				\label{tab:abla_mask_strategy}
			\end{center}
		\end{subfigure}
	}
	\hspace{.35em}
	{\begin{subfigure}{0.2\linewidth}
			\begin{center}
				\small
				\tablestyle{4pt}{1.1}
				\begin{tabular}{ccc}
					case & learnable [M] & Top-1 \\ 
					\shline
					MAE & \xmark & 83.2 \\
					MAE & \cmark & 83.6 \\
					MIM & \xmark & 83.6 \\
					\baseline{MIM} & \baseline{\cmark} & \baseline{83.7} \\
				\end{tabular}
				\caption{\textbf{Learnable mask token.}}
				\label{tab:abla_mask_token}
			\end{center}
		\end{subfigure}
	}
	\hspace{.75em}
	{\begin{subfigure}{0.4\linewidth}
			\begin{center}
				\small
				\tablestyle{4.8pt}{1.2}
				\begin{tabular}{ccccccc}
					ep.$\backslash$inp. & 64$^2$ & 96$^2$ & \baseline{128$^2$} & 160$^2$ & 192$^2$ & 224$^2$ \\ 
					\shline
					100 & 82.82 & 83.24 & 83.32 & 83.38 & 83.46 & 83.58 \\
					\baseline{400} & 83.07 & 83.51 & \baseline{83.76} & 83.78 & 83.85 & 83.93 \\
					800 & 83.23 & 83.60 & 83.84 & 83.90 & 83.96 & 84.03 \\
				\end{tabular}
				\caption{\textbf{Pre-training epoch and input resolution}. Setting: MIM, Swin-B, raw pixel as prediction target. Top-1 Acc. is reported.}
				\label{tab:abla_input_resolution}
			\end{center}
		\end{subfigure}
	}
	\\[1mm]
	{\begin{subfigure}{0.68\linewidth}
			\begin{center}
				\small
				\tablestyle{4.5pt}{1.2}
				\begin{tabular}{cccccccccc}
					enc.$\backslash$dec. & 1b256d & 1b512d & 1b768d & 4b256d & 4b512d & 4b768d & 8b256d & 8b512d & 8b768d \\ 
					\shline
					ViT-B  & \baseline{82.5} & 82.4 & 82.2 & 82.4 & 82.2 & N/A & 82.1 & 82.0 & N/A \\
					ViT-L  &  82.7 & 82.9  & 82.9  & 82.8 & 83.1 & 83.2 & 83.3 & \baseline{83.5} & 83.4 \\
					Swin-B & 83.5 & 83.5  & 83.3 & \baseline{83.6} & 83.5 & 83.3 & 83.4 & 83.2 & N/A \\
					Swin-L & 84.3 & 84.2 & 84.1 & 84.3 & \baseline{84.3} & 84.2 & 84.2 & 84.1 & 84.0 \\
				\end{tabular}
				\caption{\textbf{Decoder size}. Setting: MIM, $128^2$, HOG, 100 epochs. ``1b256d" indicates one decoder block with 256-d width. Top-1 Acc. is reported.}
				\label{tab:abla_decoder_size}
			\end{center}
		\end{subfigure}
	}
	\hspace{.2em}
	{\begin{subfigure}{0.3\linewidth}
			\begin{center}
				\small
				\tablestyle{4.5pt}{1.2}
				\begin{tabular}{ccccc}
					case & encoder & Top-1 & AP$^{\bf b}$ & AP$^{\bf m}$ \\ 
					\shline
					pixel & ViT-B & 83.8 & 50.4 & 45.0 \\
					\baseline{HOG} & \baseline{ViT-B} & \baseline{83.8} & \baseline{50.5} & \baseline{45.2} \\
					pixel & Swin-B & 84.0 & 52.3 & 46.0 \\
					\baseline{HOG} & \baseline{Swin-B} & \baseline{84.1} & \baseline{52.5} & \baseline{46.2} \\
				\end{tabular}
				\caption{\textbf{Prediction target}. Setting: MIM, 224$^2$, 800 epochs.}
				\label{tab:abla_prediction_target}
			\end{center}
		\end{subfigure}
	}
	\\[1mm]
	{\begin{subfigure}{0.61\linewidth}
			\small
			\tablestyle{4.5pt}{1.2}
			\begin{tabular}{ccccc}
				mask size$\backslash$ratio & 0.50 & 0.65 & \baseline{0.75} & 0.85 \\
				\shline
				8$\times$8 & 83.3 & 83.4 & 83.3 & 82.9 \\
				\baseline{16$\times$16} & 83.3 & 83.5 & \baseline{83.6} & 83.5 \\
				32$\times$32 & 83.6 & 83.4 & 83.2 & 83.0 \\
			\end{tabular}
			\hfill \,
			\tablestyle{4.5pt}{1.2}
			\begin{tabular}{ccccc}
				size$\backslash$ratio & 0.50 & 0.65 & \baseline{0.75} & 0.85 \\
				\shline
				16$\times$16 & 83.9 & 84.0 & 83.8 & 83.6\\
				\baseline{32$\times$32} & 83.9 & 83.9 & \baseline{84.0} & 83.8 \\
				64$\times$64\rlap{$^\dagger$} & 83.6 & 83.1 & - & - \\
			\end{tabular}
			\hfill \,
			\caption{\textbf{Mask size and mask ratio}. Setting: MIM, 128$^2$, HOG, 400 epochs, ViT-B/Swin-B. Top-1 acc is reported.}
			\label{tab:abla_mask_size_ratio}
		\end{subfigure}
	}
	\hspace{.3em}
	{\begin{subfigure}{0.36\linewidth}
			\begin{center}
				\small
				\tablestyle{4.5pt}{1.2}
				\begin{tabular}{cccccc}
					encoder & depth & Top-1 & encoder & depth & Top-1 \\ 
					\shline
					ViT-B & 8  & 82.9 & Swin-B & 22 & 83.9 \\
					\baseline{ViT-B} & \baseline{10} & \baseline{83.4} & \baseline{Swin-B} & \baseline{23} & \baseline{84.1} \\
					ViT-B & 12 & 83.5 & Swin-B & 24 & 84.1 \\
				\end{tabular}
				\caption{\textbf{Encoder depth (pre-train)}. MIM, 128$^2$, HOG, 400 epochs.}
				\label{tab:abla_encoder_depth}
			\end{center}
		\end{subfigure}
	}
	\vspace{-.5em}
	\caption{Ablation studies on several key components in MIM. We study: a-b) mask strategy; c) different pre-training epochs and input resolutions; d) decoder size; e) prediction target; f) mask size and mask ratio; g) encoder depth. ImageNet-1K top-1 accuracy, COCO box AP$^{\bf b}$ and mask AP$^{\bf m}$ are reported. (N/A: fail to converge. \dag: when the input resolution and mask size are set to 128$\times$128 and 64$\times$64, respectively, mask ratio $\textgreater$ 0.5 will lead to the same result.) Default settings are marked in \colorbox{baselinecolor}{gray}. \label{tab:ablation_study}}
	\vspace{-0.2cm}
\end{table*}

\noindent\textbf{Framework.} MIM learns representations by predicting the masked area of an input $\mathbf{X}$. Existing MIM methods can be roughly classified into two categories: (i) MAE~\cite{mae,greenmim} discards the masked area and only the visible part is sent to the encoder for latent feature extracting, then the decoder reconstructs the masked part from latent representation and mask token; (ii) SimMIM~\cite{simmim,beit,maskfeat,peco,ibot} retains the masked part, the new input can be formulated as $\mathbf{\hat{X}}=\mathbf{X}\odot\mathbf{S}+\mathbf{[M]}\odot(1-\mathbf{S})$, where $\odot$ denotes the Hadamard product.

\noindent\textbf{Encoder Architecture.} We consider two typical transformers as the encoder (backbone) for pre-training, \ie, ViT~\cite{vit} and Swin~\cite{swin}, which are both transferable to various downstream vision tasks. Therefore, the result can be directly compared with others in terms of the architecture.

\noindent\textbf{Decoder Architecture.} The latent feature extracted by encoder is then fed into the decoder, \ie, a linear layer~\cite{simmim} or several transformer blocks~\cite{mae}, to predict the original pixels~\cite{mae,simmim} or other targets~\cite{maskfeat,beit,peco} in the masked area.

\noindent\textbf{Prediction Target.} The targets can be the raw pixel values~\cite{mae}, Histograms of Oriented Gradients (HOG)~\cite{maskfeat}, context encoded via dVAE~\cite{beit,cae}, \etc.

\subsection{Empirical Observations on Basic Components}
In order to make MIM framework suitable for any vision backbones, we directly mask the original images in a block-wise manner (mask size can be adjusted in a large range), and retain all pixels during the pre-training stage. In this way, the encoder (\eg, ViT in MAE~\cite{mae} and Swin in SimMIM~\cite{simmim}) can be replaced by any architectures because the input image is of the same size as in supervised training. Here we study how the basic components influence the MIM pre-training results. All models are pre-trained on ImageNet-1K and evaluated on two benchmarks, \ie, ImageNet-1K~\cite{imagenet} and COCO~\cite{coco}, which are commonly used in previous pre-training works~\cite{mae,beit,simmim,greenmim,peco,cae}.

\noindent\textbf{Ablation 1: mask strategy.} We start out by analyzing two typical mask strategies of encoder, \ie, MAE~\cite{mae} and MIM~\cite{simmim}. The former operates only on the visible patches without [MASK] tokens, while the latter operates on the whole image patches. As shown in the top two rows of Table~\ref{tab:abla_mask_strategy}, MIM achieves slightly better transfer performance compared to MAE, but operating on whole patches will lead to a heavier computational burden (as shown in FIgure~\ref{fig:gpu_efficiency}). Besides, the third row shows that masking on image patch (\eg, 14$\times$14$\times$768 in ViT-B~\cite{vit}) has almost the same effect as masking on original image (\eg, 224$\times$224$\times$3). We further study the influence brought by the [MASK] token. Table~\ref{tab:abla_mask_token} ablates two kinds of mask token, one with learnable vector and the other is set to zeros. We find that filling mask tokens with zeros degrades the MAE by 0.4\%, but has little effect on MIM. One main reason it that the encoder in MIM can process the mask tokens earlier and more comprehensive when compared with MAE. In general, whether discarding the masked regions will not affect the final fine-tuning (transfer) result other than the pre-training computational cost.

\begin{figure*}[t]
	\centering         
	\includegraphics[width=\textwidth]{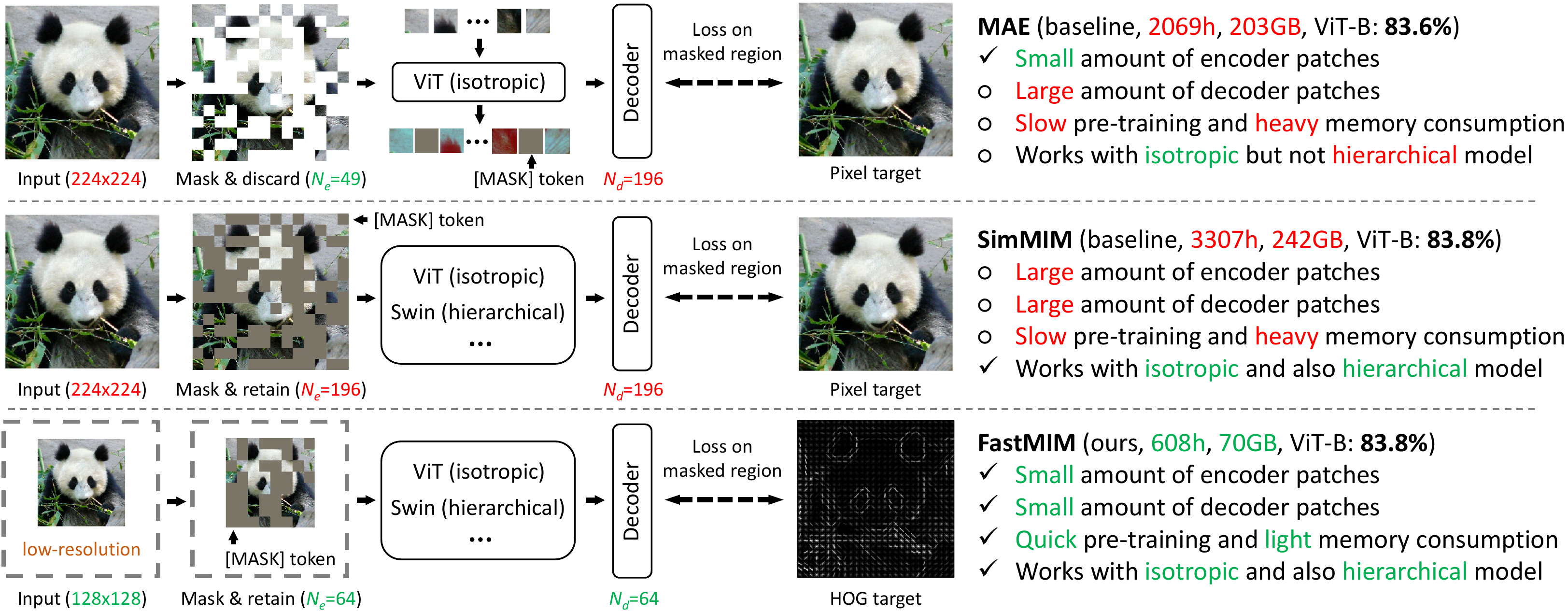}
	\vspace{-0.5cm}
	\caption{\small{Comparison among the MAE~\cite{mae}, SimMIM~\cite{simmim} and our \emph{FastMIM} framework. MAE randomly masks and discards the input patches. Although there is only small amount of encoder patches, MAE can only be used to pre-train the isotropic ViT which generates single-scale intermediate features. SimMIM preserves input resolution and can serve as a generic framework for all kinds of vision backbones, but it needs to tackle with large amount of patches. Our \emph{FastMIM} simply reduces the input resolution and replaces the pixel target with HOG target. These modifications are simple yet effective. \emph{FastMIM} (i) pre-train faster; (ii) has a lighter memory consumption; (iii) can serve as a generic framework for all kinds of architectures; and (iv) achieves comparable and even better performances compared to previous methods.}}
	\vspace{-0.2cm}
	\label{fig:fastmim}
\end{figure*}

\noindent\textbf{Ablation 2: input resolution.} Table~\ref{tab:abla_input_resolution} studies how the pre-training epoch and input resolution impact the fine-tuning result of MIM (the result of MAE can be found in supplementary material). It shows that \emph{a wide range of input resolutions} (\eg, 128$^2$ $\sim$ 224$^2$) perform equally well. The largest input resolution achieves the best top-1 accuracy as expected. It is worth noticing that when the input resolution is reduced to 128$^2$, the final fine-tuning results only drop 0.2\%/0.17\%/0.19\% in 100/400/800 epochs, respectively. More specifically, when the input resolution is set to 224$^2$, the encoder need to deal with a large number of image patches (\eg, $N_e$=56$^2$ in Swin-B stage-1), which incurs a heavy memory burden and a long computing time. In the contrast, when the input resolution is set to 128$^2$, the number of image patches naturally decreases to $N_e$=32$^2$, which is 70\% less than the former 224$^2$ input, and the result is almost at-par with it. However, we find that further reducing the input resolution results in a dramatic drop in fine-tuning top-1 accuracy, probably because lower resolution and less input patches throw too much information away, which is indispensable during the reconstruction stage. 

Table~\ref{tab:abla_input_resolution} only shows the result using pixel target. Sec.~\ref{sec:abla_on_target} presents the result using HOG target, which is more stable and achieves better performance when reducing resolution.

\noindent\textbf{Ablation 3: mask size and mask ratio.} We study how different mask sizes and mask ratios affect the effectiveness of MIM pre-training in Table~\ref{tab:abla_mask_size_ratio}. We can find that both isotropic and hierarchical architectures achieve their best results only when the mask size is equivalent to the patch size of the last stage of encoder. Notably, when the mask size is smaller than the patch size, MIM can still obtain comparable results, demonstrating its ability for representation learning. With proper mask size, MIM is quite stable with the mask ratios varying from 0.5 to 0.85.

\noindent\textbf{Ablation 4: encoder depth in pre-training.} As mentioned in ``Ablation 2", reducing the input resolution (encoder patches) can help ease memory overhead and save training time. Besides, there is another straightforward method to save computational cost: reducing the parameters (encoder depth) trained in pre-training stage. Inspired by the layer decay strategy (shallow layers have smaller learning rate compared to deep layers) in fine-tuning of BEiT~\cite{beit}, we conjecture that shallow layers are more important than deep layers during pre-training phase. As depicted in Table~\ref{tab:abla_encoder_depth}, discarding the last several layers (blocks) in pre-training stage (the discarded layers will be re-initialized in fine-tuning stage) for both ViT-B and Swin-B gains almost the same performance compared to the original setting (\eg, the third row in Table~\ref{tab:abla_encoder_depth}). Note that the hierarchical Swin-B~\cite{swin} encoder consists of four stages (\eg, [2,2,18,2]), Table~\ref{tab:abla_encoder_depth} only presents the results of [2,2,18,0] (the first row) and [2,2,18,1] (the second row). If we discard the layers in third stage, \eg, [2, 2, 16, 2], the fine-tuning performance will drop to 83.4\%.

\noindent\textbf{Ablation 5: decoder size.} Next, Table~\ref{tab:abla_decoder_size} ablates the effect of varying decoder designs. Intriguingly, the results suggest that different architectures prefer different settings and have opposite trends. ViT~\cite{vit} prefers a simple decoder for the base size and a complicated one for the large size. While Swin~\cite{swin} seems to be robust with various decoder sizes and favors a simple one, which conforms with the observation in~\cite{simmim,greenmim}. In conclusion, the size of decoder should be properly aligned with the specific encoder.

\noindent\textbf{Ablation 6: prediction target.} Table~\ref{tab:abla_prediction_target} compares the effects of two prediction targets. Predicting colors of original pixels is the most straightforward target. Specifically, we use RGB values normalized by the mean and the standard deviation of the pre-training dataset. Histograms of Oriented Gradients (HOG)~\cite{hog,maskfeat} is a feature descriptor that counts occurrences of gradient orientation in localized portions of an image. Here we minimize the $\ell_2$ distance between the model’s prediction and the ground-truth RGB value/HOG feature. Under the setting of MIM pre-training and 224$^2$ input, two prediction targets get similar performances on both classification and detection tasks. But when reducing the input resolution, pixel and HOG exhibit different characteristics, which will be further analyzed in Sec.~\ref{sec:abla_on_target}.

\section{Approach}
Our \emph{FastMIM} pre-trains vision backbones through masked image modeling, and is also a reinforced version of SimMIM~\cite{simmim}, as illustrated in the bottom of Figure~\ref{fig:fastmim}.

\subsection{Base Architecture}
In principle, it is straightforward and convenient to replace the encoder with other vision backbones in aforementioned MIM pre-training framework. We choose the representative isotropic and hierarchical vision transformers, \ie, ViT~\cite{vit} and Swin Transformer~\cite{swin} as our baselines. We directly mask the input image (\eg, $\mathbf{X} \in \mathbb{R}^{128\times 128\times 3}$) with the mask token (\eg, learnable vector [MASK] $\in \mathbb{R}^{1\times 1\times 3}$). The ViT encoder embeds patches by a linear projection added with positional embeddings (PE), while there is no extra PE for the decoder. As for Swin, the window size for Swin-B and Swin-L is set to 7 and 14 following~\cite{simmim,greenmim}, respectively. 

\subsection{\emph{\textbf{FastMIM}} Framework}
\label{sec:fastmim_framework}

\noindent\textbf{Masked Input.} The input image is randomly cropped and resized to \textbf{128$\bm\times$128}. Therefore, the number of patches (pixels) is reduced to $N_e$=64 and $N_e$=32$^2$/16$^2$/8$^2$/4$^2$ for ViT and Swin, receptively. We leverage a per-sample random mask strategy, and set the mask size to the same value as the last layer's patch size of the encoder. Specifically, the mask size is 16$\times$16 and 32$\times$32 for ViT and Swin, respectively. The mask ratio is set to 0.75 according to Table~\ref{tab:abla_mask_size_ratio}.

\noindent\textbf{Decoder.} The decoder is only used in pre-training stage to perform the reconstruction task. Note that the input resolution of \emph{FastMIM} is set to 128$\times$128, the encoder outputs for ViT-L and Swin-L are of size 64$\times$1024 and 16$\times$1536, respectively. The memory usage of our decoder is indeed 65\% less than that of MAE~\cite{mae}. According to the ablation study in Table~\ref{tab:abla_decoder_size}, the decoder sizes for ViT-B/ViT-L/Swin-B/Swin-L are set to 1b256d/8b512d/4b256d/4b512d, respectively.

\begin{table*}[t]
	\centering
	\tablestyle{5.8pt}{0.96}
	\begin{tabular}{lcccccccc}
		\toprule
		\textbf{Framework} & \textbf{Extra} & \textbf{Model} & \textbf{\# Params} & \textbf{PT Ep.} & \textbf{Hours/Ep.} & \textbf{PT Hours} & \textbf{FT Ep.} & \textbf{Top-1 (\%)} \\
		\midrule
		\multicolumn{9}{l}{\textit{Supervised pre-training}} \\
		Training from scratch in MAE~\cite{mae} & - & ViT-B & 86M & 0 & 1.6\rlap{$^{\dagger,\ddagger}$} & 490\rlap{$^{\dagger,\ddagger}$} & 300 & 82.3 \\
		Training from scratch in Swin~\cite{swin} & - & Swin-B & 88M & 0 & 2.5\rlap{$^{\dagger,\ddagger}$} & 744\rlap{$^{\dagger,\ddagger}$} & 300 & 83.5 \\
		PT \tiny({${192}$}) \footnotesize then FT \tiny({${224}$}) \footnotesize in SimMIM~\cite{simmim} & - & Swin-L & 197M & 300 & 3.8\rlap{$^\ddagger$} & 1139\rlap{$^\ddagger$} & 100 & 83.5 \\
		\midrule
		\multicolumn{9}{l}{\textit{Self-supervised pre-training with contrastive learning}} \\
		MoCov3~\cite{mocov3} & momentum ViT & ViT-B & 86M & 800 & - & - & 100 & 83.2 \\
		DINO~\cite{dino} & momentum ViT & ViT-B & 86M & 800 & - & - & 100 & 82.8 \\
		\midrule
		\multicolumn{9}{l}{\textit{Self-supervised pre-training with masked image modeling on isotropic ViT}} \\
		BEiT~\cite{beit} & DALL-E & ViT-B & 86M & 800 & 2.4 & 1920 & 100 & 83.2 \\
		MAE~\cite{mae} & - & ViT-B & 86M & 1600 & 1.3 & 2069 & 100 & 83.6 \\
		SimMIM~\cite{simmim} & - & ViT-B & 86M & 800 & 4.1 & 3307 & 100 & 83.8 \\
		LoMaR~\cite{lomar} & relative PE & ViT-B & 86M & 800 & 1.4 & 1120 & 100 & 83.8 \\
		CAE~\cite{cae} & DALL-E & ViT-B & 86M & 1600 & - & - & 100 & 83.9 \\
		MaskFeat~\cite{maskfeat} & - & ViT-B & 86M & 800 & 1.6\rlap{$^\ddagger$} & 1264\rlap{$^\ddagger$} & 100 & 84.0 \\
		iBOT~\cite{ibot} & momentum ViT & ViT-B & 86M & 1600 & - & - & 100 & 84.0 \\
		PeCo~\cite{ibot} & VQ-VAE, MoCov3 & ViT-B & 86M & 800 & - & - & 100 & 84.5 \\
		\baseline{FastMIM (ours)} & \baseline{-} & \baseline{ViT-B} & \baseline{86M} & \baseline{400} & \baseline{0.8} & \baseline{304} & \baseline{100} & \baseline{83.6} \\
		\baseline{FastMIM (ours)} & \baseline{-} & \baseline{ViT-B} & \baseline{86M} & \baseline{800} & \baseline{0.8} & \baseline{608} & \baseline{100} & \baseline{83.8} \\
		\midrule
		\multicolumn{9}{l}{\textit{Self-supervised pre-training with masked image modeling on hierarchical Swin}} \\
		SimMIM~\cite{simmim} & - & Swin-B & 88M & 800 & 2.0 & 1609 & 100 & 84.0 \\
		GreenMIM~\cite{greenmim} & - & Swin-B & 88M & 800 & 1.1 & 887 & 100 & 83.8 \\
		\baseline{FastMIM (ours)} & \baseline{-} & \baseline{Swin-B} & \baseline{88M} & \baseline{400} & \baseline{0.8} & \baseline{336} & \baseline{100} & \baseline{84.1} \\
		\midrule
		\multicolumn{9}{l}{\textit{Self-supervised pre-training with masked image modeling on hierarchical Swin}} \\
		SimMIM~\cite{simmim} & - & Swin-L & 197M & 800 & 3.5 & 2821 & 100 & 85.5\rlap{$^\ddagger$} \\
		GreenMIM~\cite{greenmim} & - & Swin-L & 197M & 800 & 1.3 & 1067 & 100 & 85.1 \\
		\baseline{FastMIM (ours)} & \baseline{-} & \baseline{Swin-L} & \baseline{197M} & \baseline{400} & \baseline{1.4} & \baseline{544} & \baseline{100} & \baseline{85.2} \\
		\baseline{FastMIM (ours)} & \baseline{-} & \baseline{Swin-L} & \baseline{197M} & \baseline{800} & \baseline{1.4} & \baseline{1088} & \baseline{100} & \baseline{85.4} \\
		\baseline{FastMIM-P (ours)} & \baseline{-} & \baseline{Swin-L} & \baseline{197M} & \baseline{400} & \baseline{1.8} & \baseline{736} & \baseline{100} & \baseline{85.5} \\
		\bottomrule
	\end{tabular}
	\vspace{-0.3cm}
	\caption{\small{Comparison with state-of-the-art MIM methods. ``PT Ep." refers to the number of pre-training epoch. ``Hours/Ep." refers to GPU hours per epoch. ``PT Hours" refers to total pre-training GPU hour. ``FT Ep." refers to fine-tuning epoch. We report top-1 accuracy on ImageNet-1K validation set with the ViT-B/Swin-B/Swin-L models. ($^\dagger$: we report the total hours in fine-tuning stage. $^\ddagger$: result tested by us.)}}
	\vspace{-0.2cm}
	\label{tab:exp_in1k}
\end{table*}

\noindent\textbf{Prediction Target.} We choose Histograms of Oriented Gradients (HOG)~\cite{hog} features as the target following MaskFeat~\cite{maskfeat}.
We first obtain an entire HOG feature on the whole image and then minimize the $\ell_2$ distance between the output of \emph{FastMIM} and original HOG feature on masked region. The number of orientation bins is set to 9, and spatial cell is set to 8$\times$8. Discussion on HOG is shown in Sec.~\ref{sec:abla_on_target}

\subsection{\emph{\textbf{FastMIM-P}}: Progressively Enlarge the Input}
To further improve the scalability of \emph{FastMIM}, we propose to progressively enlarge the input resolution during the pre-training stage, viz, \emph{FastMIM-P}. Although HOG can preserve the texture information when reducing the input resolution to some extent, the performance of model with high capacity, \eg, Swin-L, still has small gap (-0.3\% in Table~\ref{tab:exp_in1k}) compared to the counterpart trained with high-resolution input images. More specifically, in contrast to \emph{FastMIM} that trains Swin-L in fixed 128$^2$ inputs, \emph{FastMIM-P} trains Swin-L in 128$^2$/160$^2$/192$^2$ for 200/100/100 epochs (400 epochs in total), and achieve better trade-off between accuracy and training time, as shown in the last row of Table~\ref{tab:exp_in1k}.

\noindent\textbf{Discussion.} As shown in Table~\ref{tab:exp_in1k}, \emph{FastMIM-P} achieves better performance compared to \emph{FastMIM} with less pre-training time. However, as the input resolution continually increases, the GPU memory consumption will inevitably increase. The resolution and training schedule need to be carefully designed to achieve a better space-time trade-off.

\subsection{Implementation Details}
\label{sec:fastmim_exp_setup}
Following common practice~\cite{mae,simmim,beit}, we first conduct self-supervised pre-training on ImageNet-1K~\cite{imagenet} training set without label, and then validate the proposed \emph{FastMIM} by conducting end-to-end fine-tuning on downstream tasks including classification, object detection, instance segmentation, and semantic segmentation. All experiments are conducted on 8 V100 GPUs with PyTorch~\cite{pytorch}.
For the \textbf{pre-training on ImageNet-1K}, input images are transformed by simple data augmentations including random cropping and horizontal flip. MIM settings are shown in Sec.~\ref{sec:fastmim_framework}. All models are trained with AdamW~\cite{adamw} optimizer under cosine annealing schedule. We set the base learning rate (\emph{lr}) to 1.5e-4, and the effective \emph{lr} is scaled linearly.
For the \textbf{fine-tuning on ImageNet-1K}, each backbone is fine-tuned for 100 epochs with strong data augmentation including label smoothing, mixup, and cutmix following~\cite{mae,simmim}. The drop path rates are set to 0.1/0.1/0.3 for ViT-B/Swin-B/Swin-L, respectively. Notably, we report the best top-1 accuracy through the \textbf{\emph{grid search}} on base \emph{lr} and layer-wise \emph{lr} decay rate. 
For the \textbf{fine-tuning on COCO}, we choose typical Mask R-CNN~\cite{maskrcnn} as the basic framework for object detection and instance segmentation tasks. For ViT, we follow the training setup in MAE~\cite{mae}. For Swin Transformer, we follow two training setups in both SimMIM~\cite{simmim} and GreenMIM~\cite{greenmim} for fair comparison.  
For the \textbf{fine-tuning on ADE20K}, we use UperNet~\cite{upernet} and follow the settings in both BEiT~\cite{beit} and SimMIM~\cite{simmim} for fair comparison. More implementation details can be found in the appendix.

\section{Experiments}

\subsection{ImageNet-1K Classification} Table~\ref{tab:exp_in1k} reports the top-1 accuracy on ImageNet validation set~\cite{imagenet}. We compare our \emph{FastMIM} with vision transformers trained via supervised pre-training, self-supervised training with contrastive learning, and self-supervised training with MIM. Compared to the models trained from scratch with random initialization, we find that pre-training through \emph{FastMIM} significantly improves performances on both ViT-B and Swin-B by +1.5\% and +0.6\%, respectively. Notably, the total pre-training hour of \emph{FastMIM} based Swin-B is 336, the total pre-training and fine-tuning time is 584 hours, which is \emph{less} than the 744 hours for training from scratch. This result indicates that our \emph{FastMIM} can also serve as a regular training paradigm for classification, and is more efficient and effective than the commonly used scheme. Besides, ViT-B pre-trained with 400 epochs via \emph{FastMIM} achieves 83.6\% top-1 accuracy on ImageNet-1K, which is +1.3\% higher than the baseline counterpart. And the total pre-training and fine-tuning time is 467 hours, which is also \emph{less} than 490 hours spent by conventional supervised scheme. These improvements suggest that our \emph{FastMIM} can effectively expedite the pre-training process for various vision backbones.

Moreover, we compare \emph{FastMIM} with previous state-of-the-art self-supervised methods for isotropic ViT-B, such as BEiT~\cite{beit}, MAE~\cite{mae}, SimMIM~\cite{simmim}, CAE~\cite{cae}, MaskFeat~\cite{maskfeat}, iBOT~\cite{ibot}, and PeCo~\cite{peco}. Among them, CAE~\cite{cae} uses extra 250M DALL-E data~\cite{dalle} to pre-train the tokenizer, iBOT~\cite{ibot} uses an extra momentum ViT as the online tokenizer, and PeCo~\cite{peco} leverages both VQ-VAE~\cite{vqvae} tokenizer and MoCov3~\cite{mocov3} framework. These extra modules introduce non-negligible memory overhead and considerably longer training time. MAE~\cite{mae}, SimMIM~\cite{simmim} and MaskFeat~\cite{maskfeat} are the most comparable methods. Our approach achieves 83.8\% top-1 accuracy, which is on par with above MIM frameworks. As for the computational cost, \emph{FastMIM} is 1.6$\times$/2$\times$/5.1$\times$ faster than MAE/SimMIM/MaskFeat, and reduces the GPU memory consumption by 50\%$\sim$70\% when compared to MAE and SimMIM, as shown in Figure~\ref{fig:gpu_efficiency}.

\begin{table*}[t]
	\small
	\centering
	\tablestyle{6.5pt}{0.98}
	\begin{tabular}{lccccccccccc}
		\toprule
		\multirow{2}{*}{\textbf{Framework}} & \multirow{2}{*}{\textbf{Backbone}} & \multirow{2}{*}{\textbf{IN-1K FT}} & \multirow{2}{*}{\textbf{PT Epoch}} & \multirow{2}{*}{\textbf{PT Hours}} & \multirow{2}{*}{\textbf{FT Epoch}} & \multicolumn{3}{c}{{Object Detection}} & \multicolumn{3}{c}{{Instance Segmentation}} \\
		& & & & & & \textbf{AP$^{\mathrm{b}}$} & \textbf{AP$_{50}^{\mathrm{b}}$} & \textbf{AP$_{75}^{\mathrm{b}}$} & \textbf{AP$^{\mathrm{m}}$} & \textbf{AP$_{50}^{\mathrm{m}}$} & \textbf{AP$_{75}^{\mathrm{m}}$} \\
		\midrule
		\multicolumn{12}{l}{\textit{Training from scratch (random initialization)}} \\
		Benchmarking~\cite{benchmarking} & ViT-B & \xmark & 0 & 0 & 400 & 48.9 & - & - & 43.6 & - & - \\
		\midrule
		\multicolumn{12}{l}{\textit{Self-supervised pre-training, follow the coco fine-tuning setup in MAE~\cite{mae,benchmarking}}} \\
		{MAE~\cite{mae}} & ViT-B & \xmark & 1600 & 2069 & 25 & 48.1 & 69.3 & 53.3 & 43.2 & 66.3 & 46.7 \\
		\baseline{FastMIM (ours)} & \baseline{ViT-B} & \baseline{\xmark} & \baseline{800} & \baseline{608} & \baseline{25} & \baseline{48.6} & \baseline{70.5} & \baseline{53.6} & \baseline{43.5} & \baseline{67.0} & \baseline{46.9} \\
		BEiT~\cite{beit} & ViT-B & \xmark & 800 & 1920 & 100 & 49.8 & - & - & 44.4 & - & - \\
		{MAE~\cite{mae}} & ViT-B & \xmark & 1600 & 2069 & 100 & 50.3 & 70.9 & 55.6 & 44.9 & 68.3 & 49.0 \\
		\baseline{FastMIM (ours)} & \baseline{ViT-B} & \baseline{\xmark} & \baseline{800} & \baseline{608} & \baseline{100} & \baseline{50.7} & \baseline{71.3} & \baseline{56.0} & \baseline{45.1} & \baseline{68.6} & \baseline{49.3} \\
		\midrule
		\multicolumn{12}{l}{\textit{Self-supervised pre-training, follow the coco fine-tuning setup in GreenMIM~\cite{greenmim}}} \\
		SimMIM~\cite{simmim} & Swin-B & \xmark & 800 & 1609 & 36 & 50.4 & 70.9 & 55.5 & 44.4 & 68.2 & 47.9 \\
		GreenMIM~\cite{greenmim} & Swin-B & \xmark & 800 & 887 & 36 & 50.0 & 70.7 & 55.4 & 44.1 & 67.9 & 47.5 \\
		\baseline{FastMIM (ours)} & \baseline{Swin-B} & \baseline{\xmark} & \baseline{400} & \baseline{336} & \baseline{36} & \baseline{50.3} & \baseline{71.0} & \baseline{55.3} & \baseline{44.4} & \baseline{68.2} & \baseline{48.0} \\
		\midrule
		\multicolumn{12}{l}{\textit{Self-supervised pre-training, follow the coco fine-tuning setup in SimMIM~\cite{simmim}}} \\
		SimMIM{$^\dagger$}~\cite{simmim} & Swin-B & \cmark & 800 & 1609 & 36 & 52.3 & 73.4 & 57.9 & 46.1 & 70.6 & 50.2 \\
		\baseline{FastMIM (ours)} & \baseline{Swin-B} & \baseline{\xmark} & \baseline{400} & \baseline{336} & \baseline{36} & \baseline{51.9} & \baseline{72.9} & \baseline{57.2} & \baseline{45.8} & \baseline{70.2} & \baseline{49.5} \\
		\baseline{FastMIM (ours)} & \baseline{Swin-B} & \baseline{\cmark} & \baseline{400} & \baseline{336} & \baseline{36} & \baseline{52.2} & \baseline{73.3} & \baseline{57.6} & \baseline{46.1} & \baseline{70.4} & \baseline{50.2} \\
		SimMIM{$^\dagger$}~\cite{simmim} & Swin-L & \cmark & 800 & 2821 & 36 & 53.7 & 74.8 & 58.6 & 47.2 & 71.9 & 51.5 \\
		\baseline{FastMIM (ours)} & \baseline{Swin-L} & \baseline{\cmark} & \baseline{400} & \baseline{544} & \baseline{36} & \baseline{53.2} & \baseline{74.4} & \baseline{58.1} & \baseline{46.9} & \baseline{71.6} & \baseline{51.3} \\
		\baseline{FastMIM-P (ours)} & \baseline{Swin-L} & \baseline{\cmark} & \baseline{400} & \baseline{736} & \baseline{36} & \baseline{53.6} & \baseline{74.9} & \baseline{58.4} & \baseline{47.2} & \baseline{72.0} & \baseline{51.5} \\
		\bottomrule
	\end{tabular}
	\vspace{-0.2cm}
	\caption{\small{COCO object detection and instance segmentation. All methods are based on the Mask R-CNN~\cite{maskrcnn} architecture with the FPN neck. ``IN-1K FT" indicates whether use the model fine-tuned on ImageNet-1K for the initialization on COCO. ($^\dagger$: our implementation, the IN-1K fine-tuned checkpoint is downloaded from github, and the final AP$^{\mathrm{b}}$ is similar with the number reported in SimMIM paper~\cite{simmim}.)}}
	\vspace{-0.3cm}
	\label{tab:exp_coco}
\end{table*}

\begin{table}[t]
	\small
	\centering
	\tablestyle{4.pt}{0.98}
	\begin{tabular}{lcccc}
		\toprule
		\textbf{Framework} & \textbf{Backbone} & \textbf{PT Epoch} & \textbf{PT Hours} & \textbf{mIoU} \\
		\midrule
		\multicolumn{5}{l}{\textit{Self-supervised pre-training, follow the setup in MAE~\cite{mae}}} \\
		MoCov3~\cite{mocov3} & ViT-B & - & - & 47.3 \\
		BEiT (w/ DALL-E)~\cite{beit} & ViT-B & 800 & 1920 & 47.1 \\
		MAE~\cite{mae} & ViT-B & 1600 & 2069 & 48.1 \\
		PeCo~\cite{peco} & ViT-B & 800 & - & 48.5 \\
		CAE (w/ DALL-E)~\cite{cae} & ViT-B & 800 & - & 49.7 \\
		\baseline{FastMIM (ours)} & \baseline{ViT-B} & \baseline{800} & \baseline{608} & \baseline{49.4} \\
		\midrule
		\multicolumn{5}{l}{\textit{Self-supervised pre-training, follow the setup in SimMIM~\cite{simmim}}} \\
		SimMIM~\cite{simmim} & Swin-B & 800 & 1609 & 52.8 \\
		\baseline{FastMIM (ours)} & \baseline{Swin-B} & \baseline{400} & \baseline{336} & \baseline{52.6} \\
		\bottomrule
	\end{tabular}
	\vspace{-0.2cm}
	\caption{\small{Semantic segmentation on ADE20K. We report the results of ViT-B and Swin-B following two settings.}}
	\vspace{-0.4cm}
	\label{tab:exp_ade20k}
\end{table}

In addition, we evaluate \emph{FastMIM} with hierarchical Swin Transformer~\cite{swin}.
Our approach obtains 84.1\% top-1 accuracy with the Swin-B backbone, which is superior to the supervised learning counterpart. When compared to the recently proposed GreenMIM~\cite{greenmim}, which exclusively designs a group window attention for pre-training Swin, \emph{FastMIM} achieves slightly better result (+0.3\%) with only half of the pre-training time, and less memory usage (108.3 \vs 121.6). And our \emph{FastMIM} can serve as a generic MIM framework for various architectures such as ViT~\cite{vit}, Swin~\cite{swin}, PVT~\cite{pvt}, CMT~\cite{pvt}, Twins~\cite{twins}. (More results in appendix.)

As for Swin-L, we fine-tune the SimMIM~\cite{simmim} through the grid search, as mentioned in Sec.~\ref{sec:fastmim_exp_setup}, the result is slightly better than that reported in SimMIM paper. When pre-trained with 400 epochs, \emph{FastMIM} achieves 85.2\% top-1 accuracy and surpasses the 800 epochs GreenMIM~\cite{greenmim}. When the pre-training epoch is extend to 800, \emph{FastMIM} further improves the Swin-L by +0.2\%. Besides, \emph{FastMIM-P} achieves 85.5\% top-1 accuracy, which is at-par with the result obtained by SimMIM trained with 800 epochs, while our pre-training speed is $\sim$4$\times$ faster. The corresponding results demonstrate the effectiveness and efficiency of our method, especially the substantial improvements on pre-training speed and memory consumption over previous MIM frameworks.

\subsection{Object Detection and Instance Segmentation}
We show the transfer learning results on COCO~\cite{coco} in Table~\ref{tab:exp_coco}. We first follow the fine-tuning setting in MAE~\cite{mae,benchmarking}, and report results of two considered training lengths: 25 and 100 epochs. Our \emph{FastMIM} yields up to 0.5 and 0.4 higher AP$^{\mathrm{box}}$ than MAE in both settings, and the pre-training hours is much less than MAE. Then we directly use the code base of the GreenMIM~\cite{greenmim} without any modification to the fine-tuning strategy. Compared with the Swin-B pre-trained by GreenMIM, our approach performs prominently better in terms of all metrics, \eg, +0.3\% improvement in both AP$^{\mathrm{box}}$ and AP$^{\mathrm{mask}}$, with less pre-training epochs (-400). Besides, our approach still obtains similar results with the SimMIM~\cite{simmim}. Our \emph{FastMIM} can also scale up to larger models and obtain better performance. We conduct the experiments by following the settings in SimMIM, and use their public checkpoints for direct comparisons. Our \emph{FastMIM} achieves 52.2 and 53.2 AP$^{\mathrm{box}}$ (46.1 and 46.9 AP$^{\mathrm{mask}}$) for Swin-B and Swin-L, respectively, which are comparable to the SimMIM, and are achieved with much less pre-training cost. Furthermore, \emph{FastMIM-P} obtains almost the same performance as SimMIM with faster pre-training speed. In general, the masked image modeling based methods show the potential to substantially improve detection transfer learning results, and our \emph{FastMIM} can save a lot of pre-training overhead and bring impressive pre-training efficiency.

\subsection{ADE20K Semantic Segmentation}
Table~\ref{tab:exp_ade20k} presents the result of \emph{FastMIM} on ADE20K~\cite{ade20k}. Following the setup in MAE~\cite{mae}, we achieve 49.4 mIoU, +1.3 better than MAE while requiring only 30\% of its pre-training time. We note that the performance is also comparable to CAE~\cite{cae}, which leverages extra DALL-E data to pre-train its tokenizer. Besides, we follow the setup in SimMIM~\cite{simmim} and obtain 52.6 mIoU, which is also comparable to the 52.8 obtained by SimMIM.

\subsection{Ablation on Epoch/Resolution/Target}
\label{sec:abla_on_target}
\begin{figure}[t]
	\centering
	{\begin{subfigure}{\linewidth}
			\centering         
			\includegraphics[width=\linewidth]{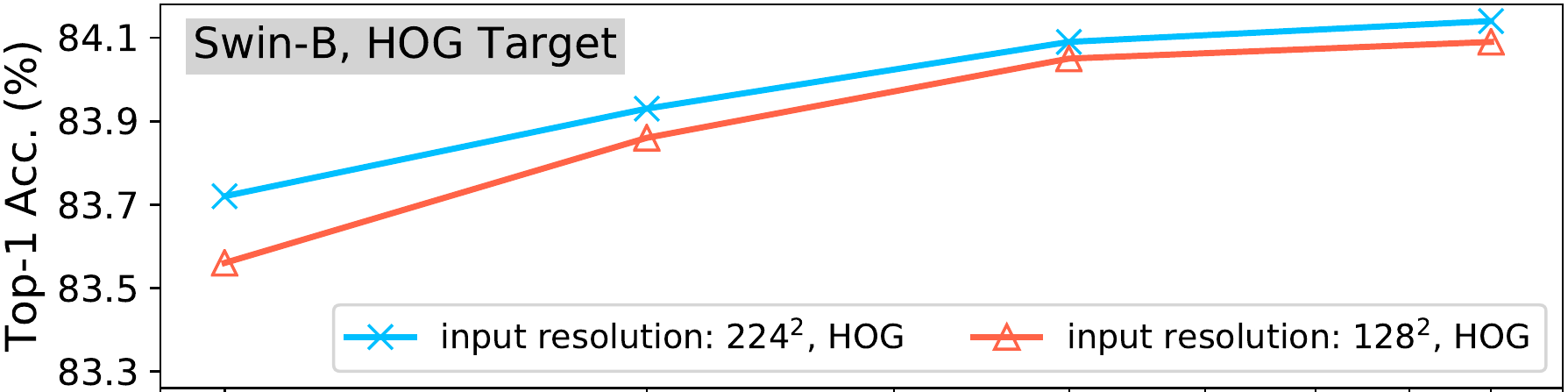}
			\\[0.5mm]
			\includegraphics[width=\linewidth]{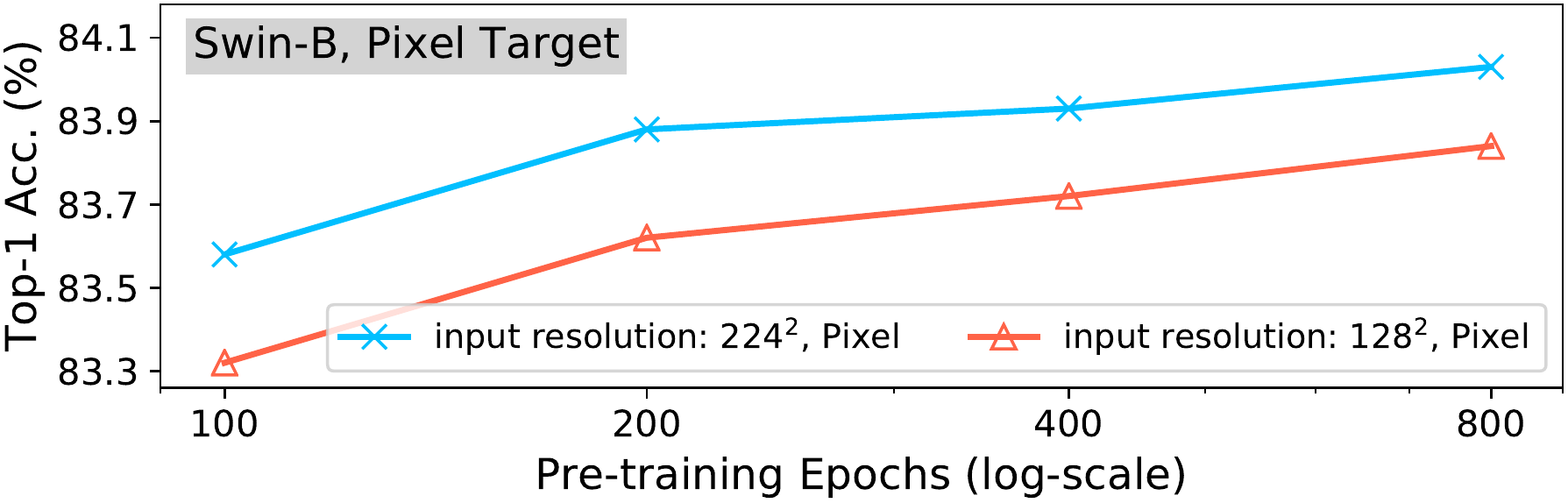}
			\vspace{-0.4cm}
		\end{subfigure}
	}
	\vspace{-0.3cm}
	\caption{\small{Ablation on training epoch/resolution/target. A longer epoch gives a noticeable improvement. HOG can preserve better performance under low resolution input compared to pixel target.}}
	\label{fig:abla_swin_hog_pixel}
	\vspace{-0.3cm}
\end{figure}

\begin{figure}[t]
	\centering
	{\begin{subfigure}{\linewidth}
			\centering         
			\includegraphics[width=\linewidth]{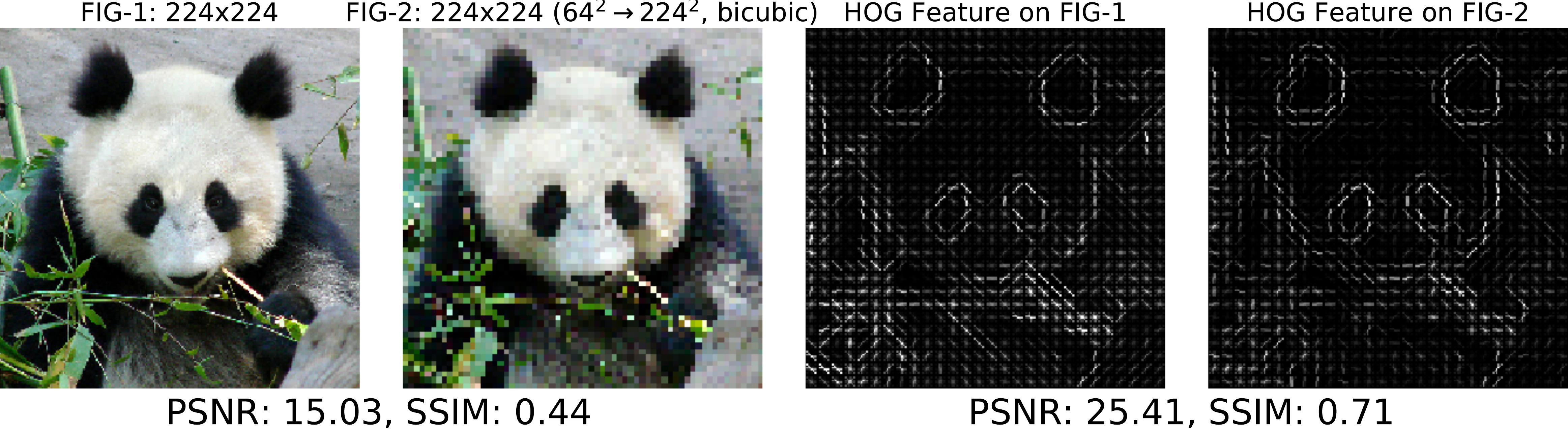}
			\\[0.5mm]
			\includegraphics[width=\linewidth]{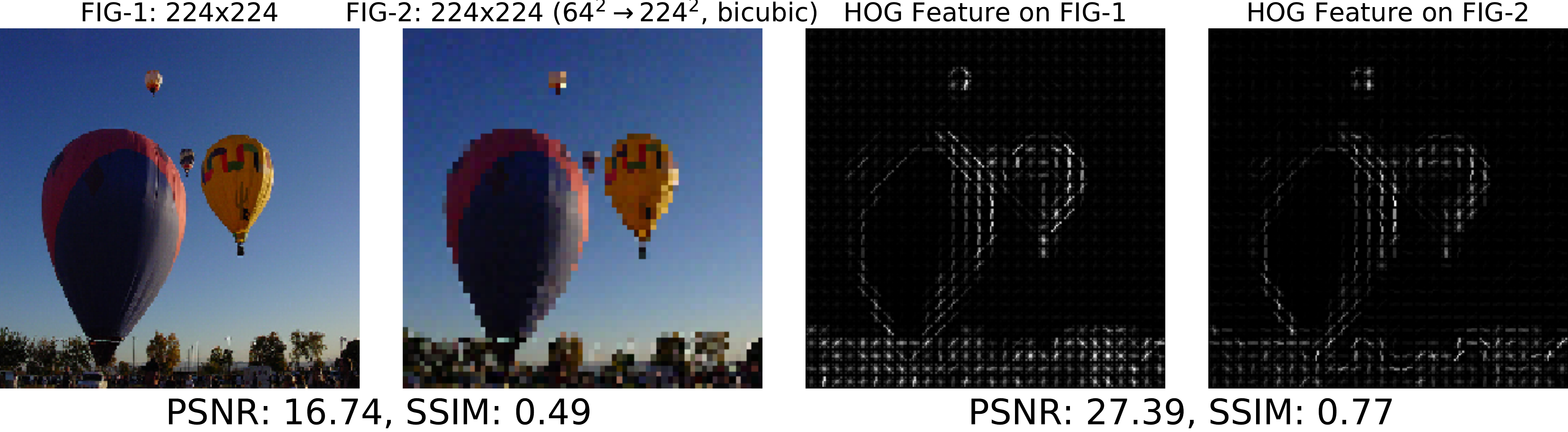}
			\\[0.5mm]
			\includegraphics[width=\linewidth]{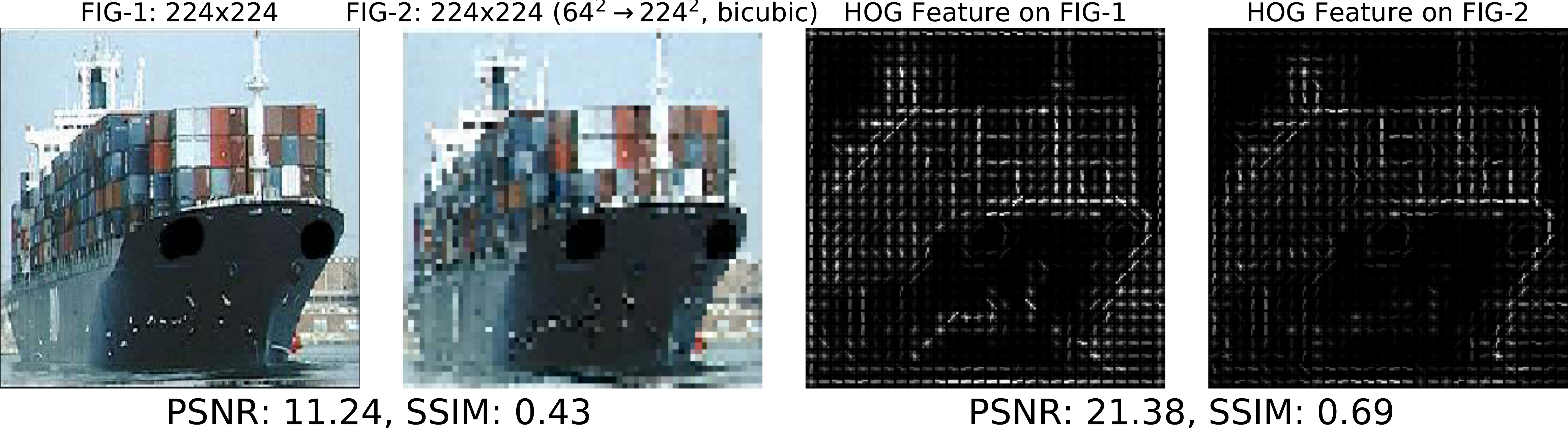}
			\vspace{-0.4cm}
		\end{subfigure}
	}
	\vspace{-0.3cm}
	\caption{\small{Visualization on pixel target and HOG target. We choose PSNR(dB) and SSIM~\cite{psnr} to evaluate the similarity between two images (features). HOG target can preserve better texture information under low resolution input compared to pixel target.}}
	\label{fig:vis_swin_hog_pixel}
	\vspace{-.1in}
\end{figure}

Figure~\ref{fig:abla_swin_hog_pixel} shows the influence of training schedule length, input resolution, and prediction target. The accuracy improves steadily as training epochs increase. We observe that HOG target starts to saturate at 400 epochs, and this behavior is contrast to pixel target. One main reason is that HOG is more robust to ambiguity by histogramming local gradients~\cite{maskfeat}. Besides, HOG can preserve better performance when reducing the input resolution, attributed to its characteristic. To elaborate it, we first qualitatively compare HOG to pixel as the prediction target in Figure~\ref{fig:vis_swin_hog_pixel}. Although reducing the image resolution can significantly expedite the training process, the crucial information, \eg, detailed textures and edges, will be discarded when using pixel target. However, HOG is more invariant to the resolution changes, which is suitable for our \emph{FastMIM}. In addition, we also show the values of pre-training loss in Table~\ref{tab:loss_hog_pixel}. Obviously, HOG can reduce the gap of loss values of different resolutions. And the loss of using HOG target is far smaller than that of using pixel target, demonstrating that HOG can effectively reduce the risk of ambiguity during reconstruction in MIM.

\section{Related Work}

\noindent\textbf{Masked Image Modeling.}
Motivated by tremendously successful BERT~\cite{bert} and its variants~\cite{brown2020language} for MLM in NLP field, masked image modeling (MIM) is first studied in BEiT~\cite{beit} to pre-train vision transformers~\cite{vit}. BEiT randomly masks a portion of image patches, and adopts a VQ-VAE~\cite{vqvae} as the visual tokenizer to generate reconstruction targets to finally predict the visual tokens which are corresponding to the masked regions. Recently, several works~\cite{simmim,ibot,peco,mae,greenmim,cae,maskfeat} have revisited MIM as a promising solution to visual representation learning. MAE~\cite{mae} develops an asymmetric encoder-decoder architecture, with an encoder that operates only on the visible patches (discarding the masked patches), along with a lightweight decoder that reconstructs the masked patches. However, MAE can only be applied to isotropic backbones. In contrast, SimMIM~\cite{simmim} proposes to retain all input patches~\cite{peco,beit,ibot,lomar} and thus can serve as generic MIM approach for hierarchical backbones. However, the large amount of input patches not only slow down its pre-training speed, but also incur heavy memory consumption, making SimMIM hard to be deployed on single deep learning machine.

\begin{table}[t]
	\centering
	\small
	\tablestyle{6pt}{0.9}
	\begin{tabular}{lcccc}
		\toprule
		\multirow{2}{*}{\textbf{Model}} & \multicolumn{2}{c}{HOG Target} & \multicolumn{2}{c}{Pixel Target} \\
		& \textbf{Input} & \textbf{Loss} & \textbf{Input} & \textbf{Loss} \\
		\midrule
		ViT-B  & 224$^2$/128$^2$ & 0.028/0.031 & 224$^2$/128$^2$ & 0.408/0.494 \\
		Swin-B & 192$^2$/128$^2$ & 0.034/0.037 & 192$^2$/128$^2$ & 0.521/0.619 \\
		Swin-L & 192$^2$/128$^2$ & 0.031/0.035 & 192$^2$/128$^2$ & 0.514/0.594 \\
		\bottomrule
	\end{tabular}
	\vspace{-0.2cm}
	\caption{\small{Ablation on the value of pre-training loss. ViT-B is trained with 800 epochs, Swin-B and Swin-L are trained with 400 epochs.}}
	\vspace{-0.3cm}
	\label{tab:loss_hog_pixel}
\end{table}

\noindent\textbf{Expedite MIM.} An obstacle for practical applications of above MIM is the heavy computational cost and long pre-training time. Towards this, UM-MAE~\cite{ummae} designs a secondary masking strategy to preserve equivalent elements across multiple local window. LoMaR~\cite{lomar} performs masked reconstruction within a small window of 7$\times$7 patches. GreenMIM~\cite{greenmim} proposes a group window attention exclusively for hierarchical Swin Transformer~\cite{swin}. In contrast to them, our \emph{FastMIM} directly reduce the input resolution, introducing no additional modification to encoder compared with supervised training paradigm, and achieves better trade-off between pre-training speed and fine-tuning accuracy.

\noindent\textbf{Reconstruction Target in MIM.}
In addition to discrete tokens~\cite{beit,peco} mentioned above, there are still various target signals designed for MIM, such as normalized pixels~\cite{mae,simmim}, HOG~\cite{maskfeat}, and latent features~\cite{ibot,data2vec}. Among them, pixel and HOG can be directly obtained from original input without extra trained networks. The histogram of oriented gradients (HOG) is a feature descriptor that counts occurrences of gradient orientation in localized portions of an image. And we demonstrate that HOG target is more invariant to the geometric changes in input image and preserves better performance under low resolution input compared to pixel target.

\section{Conclusion}

This paper presents a simple yet effective \emph{FastMIM} to expedite the self-supervised MIM pre-training for various vision backbones. As a generic framework, we directly mask input images, allowing all encoders to be trained in the same way as supervised learning. Besides, simply reducing the image resolution and reconstructing HOG target can train both isotropic and hierarchical architectures 5$\times$ faster and save the GPU memory consumption by up to $\sim$70\% compared with previous approaches, while obtaining a comparable performance on classification and other downstream vision tasks. We hope our empirical observations and the simple framework can make MIM more practicable and demolish the barrier so that more researchers can dive into this field.

\appendix

\section*{Appendix}
In this supplementary material, we first list the detailed experimental settings for pre-training, fine-tuning, and transfer learning on downstream tasks in Sec.~\ref{appendix_sec:impl_detail}. Then we show more classification results about \emph{FastMIM} for other vision backbones in Sec.~\ref{appendix_sec:exp_other_bb}. We further explore the impact of reducing input resolution in MAE~\cite{mae} in Sec.~\ref{appendix_sec:reduce_mae}. Finally we present more ablation studies together with visualization results about pixel target and HOG target in Sec.~\ref{appendix_sec:abla_hog}.

\section{Implementation Details}
\label{appendix_sec:impl_detail}
\subsection{ImageNet Experiments}
\noindent\textbf{ViT architecture.} We follow the standard ViT architecture~\cite{vit}. The encoder ends with an extra Layer Normalization (LN)~\cite{ln}. To match the different widths between encoder and decoder, we adopt a linear projection layer after the encoder following~\cite{mae}. Our \emph{FastMIM} only adds absolute positional embeddings (the sine-cosine version) to the encoder inputs. And we retain the class token~\cite{vit} during our pre-training stage.

\noindent\textbf{Swin architecture.} We follow the standard Swin-B architecture~\cite{swin}. When pre-training with input images of size 128$\times$128, the window size is set to 4 accordingly. When fine-tuning with input images of size 224$\times$224, the window size is set to 7. And we simply leverage the ``bicubic" interpolation to remap ``relative position table"~\cite{swin} when pre-trained window size mismatches with fine-tuned window size. As for Swin-L, we set the window size to 14 during fine-tuning stage following~\cite{simmim,greenmim}. \emph{FastMIM} pre-trains Swin-L with 128$\times$128 inputs and the window size is set to 8 accordingly. \emph{FastMIM-P} gradually increases the input resolution during pre-training stage. We first initialize the window size to 14, and then interpolate the corresponding ``relative position table" for different input resolutions.

\noindent\textbf{Pre-training.} The default setting is shown in Table~\ref{tab:impl_pretrain}. We simply use random resized cropping for data augmentation. We follow the official codes of ViT~\cite{vit} and Swin~\cite{swin} to initialize corresponding blocks. We set the base learning rate to 1.5e-4, and the effective learning rate is scaled linearly: $lr$ = $base\_lr \times batch\_size~/~256$.

\begin{table}[t]
	\centering
	\small
	\setlength{\tabcolsep}{5pt}
	\begin{tabular}{y{87}|x{130}}
		config & ViT-B~\cite{vit}, Swin-B~\cite{swin}, Swin-L~\cite{swin} \\
		\shline
		optimizer & AdamW \cite{adamw} \\
		base learning rate & 1.5e-4 \\
		weight decay & 0.05 \\
		optimizer momentum & $\beta_1, \beta_2{=}0.9, 0.95$ \cite{chen2020generative} \\
		batch size & 2048 \\
		learning rate schedule & cosine~\cite{sgdr}, cosine~\cite{sgdr}, step~\cite{simmim} \\
		warmup epochs & 10 \\
		pre-training epochs & 800, 400, 400 \\
		augmentation & RandomResizedCrop \\
	\end{tabular}
	\vspace{-.5em}
	\caption{\small{Hyperparameters for pre-training ViT-B, Swin-B, and Swin-L on ImageNet-1K.}}
	\label{tab:impl_pretrain}
	\vspace{-.5em}
\end{table}

\begin{table}[t]
	\centering
	\small
	\setlength{\tabcolsep}{5pt}
	\begin{tabular}{y{87}|x{130}}
		config & ViT-B~\cite{vit}, Swin-B~\cite{swin}, Swin-L~\cite{swin} \\
		\shline
		optimizer & AdamW \cite{adamw} \\
		base learning rate & 1.0e-3 \\
		weight decay & 0.05 \\
		optimizer momentum & $\beta_1, \beta_2{=}0.9, 0.999$ \cite{chen2020generative} \\
		layer-wise lr decay~\cite{lrdecay,beit} & 0.7, 0.8, 0.75\\
		batch size & 1024 \\
		learning rate schedule & cosine~\cite{sgdr} \\
		warmup epochs & 5 \\
		training epochs & 100 \\
		augmentation & RandAug (9, 0.5)~\cite{randaugment} \\
		label smoothing~\cite{labelsmooth} & 0.1 \\
		mixup~\cite{mixup} & 0.8 \\
		cutmix~\cite{cutmix} & 1.0 \\
		drop path rate~\cite{dpr} & 0.1, 0.1, 0.3 \\
	\end{tabular}
	\vspace{-.5em}
	\caption{\small{Hyperparameters for fine-tuning ViT-B, Swin-B, and Swin-L on ImageNet-1K.}}
	\label{tab:impl_ft_imagenet}
	\vspace{-.5em}
\end{table}

\noindent\textbf{Fine-tuning on ImageNet-1K.} The default setting is shown in Table~\ref{tab:impl_ft_imagenet}. We follow previous practice~\cite{beit,mae} and use a layer-wise learning rate decay strategy~\cite{lrdecay,beit} for fine-tuning. We fine-tune each backbone for 100 epochs with strong data augmentation including label
smoothing~\cite{labelsmooth}, mixup~\cite{mixup}, and cutmix~\cite{cutmix} following MAE~\cite{mae} and SimMIM~\cite{simmim}. The drop path rates~\cite{dpr} are set to 0.1/0.1/0.3 for ViT-B/Swin-B/Swin-L, respectively. To be noticed, we report the best top-1 accuracy through the grid search on base learning rate and layer-wise learning rate decay, as discussed in Sec.~\ref{sec:grid_search} .

\begin{figure}[t]
	\centering
	{\begin{subfigure}{\linewidth}
			\centering         
			\includegraphics[width=\linewidth]{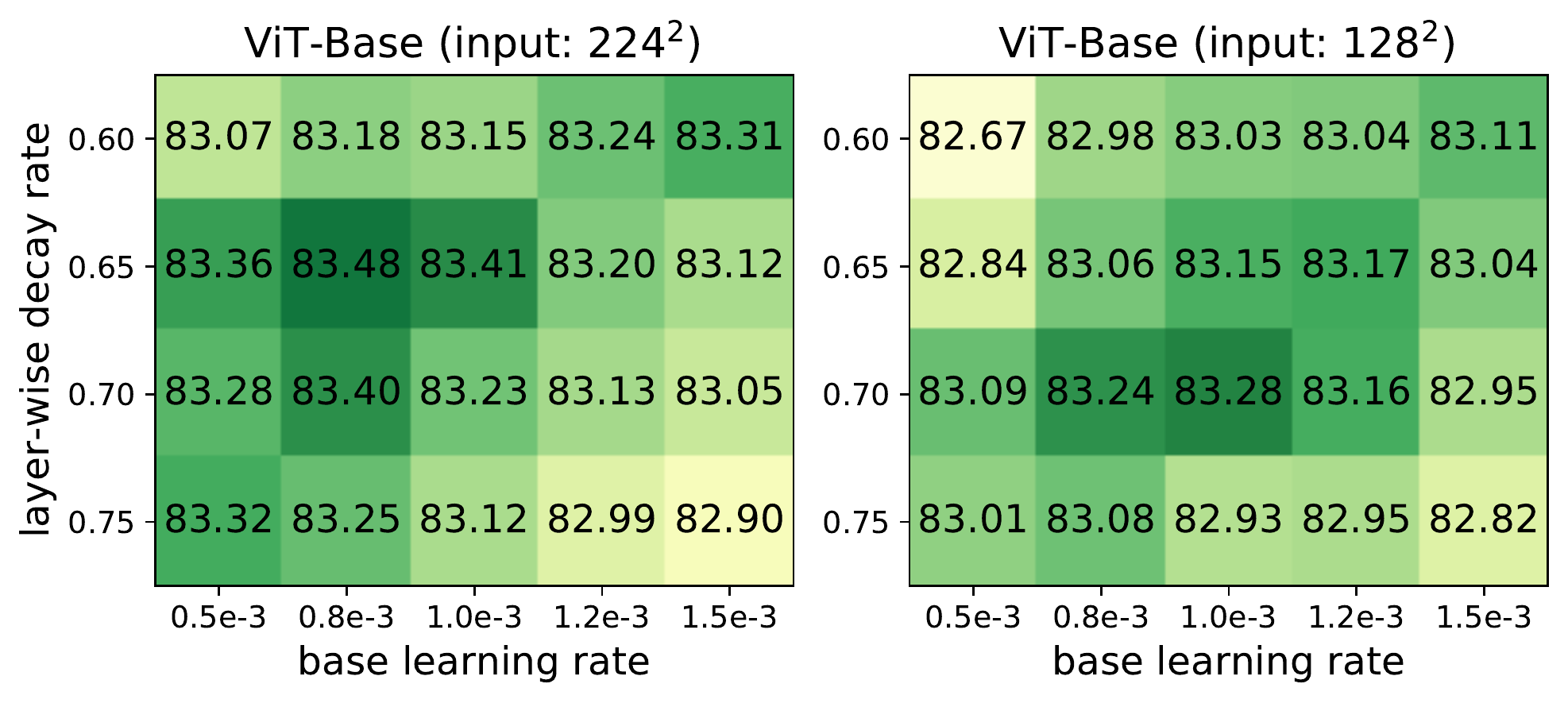}
			\\[0.5mm]
			\includegraphics[width=\linewidth]{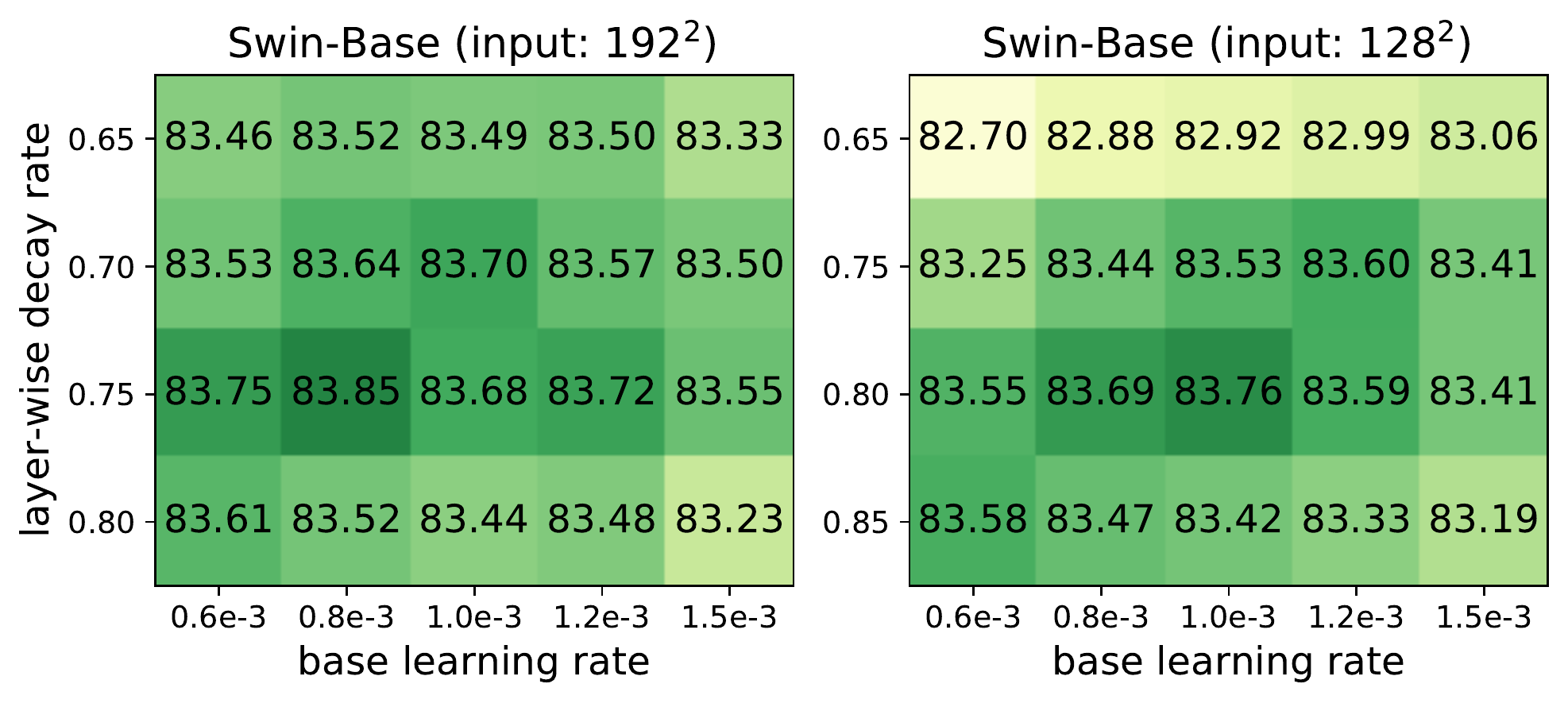}
			\vspace{-0.4cm}
		\end{subfigure}
	}
	\vspace{-0.2cm}
	\caption{\small{Grid search for fine-tuning hyperparameters. Top: ViT-B pre-trained with 400 epochs and pixel target. Bottom: Swin-B pre-trained with 400 epochs and pixel target. Deeper color indicates higher top-1 accuracy on ImageNet-1K validation set.}}
	\label{supp_fig:grid_search}
	\vspace{-0.2cm}
\end{figure}

\subsection{Hyperparameters for Fine-tuning}
\label{sec:grid_search}
To better adapt the pre-training formula to each model, we carefully sweep two hyperparameters via grid search in fine-tuning stage: (i) base learning rate (\emph{blr}), and (ii) layer-wise decay rate (\emph{ldr}), while keeping all others the same for all models. We conducted pilot experiments using ViT-B~\cite{vit} and Swin-B~\cite{swin} pre-trained with our \emph{FastMIM} to estimate reasonable hyperparameter ranges. We center a 3$\times$3 grid at \emph{blr}, \emph{ldr} = \{1.0e-3, 0.75\} and use larger and smaller values around the center. If a local optimum is not found, \ie, the best value is a boundary value, we expand the search. Figure~\ref{supp_fig:grid_search} shows corresponding results of ViT-B and Swin-B.

\subsection{Object Detection and Segmentation on COCO}
We adapt the ViT and Swin for the use of an FPN backbone~\cite{fpn} in Mask R-CNN~\cite{maskrcnn}. We follow three commonly used settings for fair comparison with other methods.

\begin{table}[!ht]
	\centering
	\small
	\setlength{\tabcolsep}{5pt}
	\begin{tabular}{y{87}|x{130}}
		config & ViT-B~\cite{vit}, Swin-B~\cite{swin}, Swin-L~\cite{swin} \\
		\shline
		optimizer & AdamW \cite{adamw} \\
		peak learning rate & 8e-5, 6e-5, 6e-5 \\
		weight decay & 0.1, 0.05, 0.05 \\
		optimizer momentum & $\beta_1, \beta_2{=}0.9, 0.999$ \cite{chen2020generative} \\
		batch size & 32 \\
		learning rate schedule & cosine~\cite{sgdr}, step, step \\
		warmup steps & 1500 \\
		training epochs & 25 \& 100, 36, 36 \\
		input resolution & (1024, 1024) \\
		drop path rate~\cite{dpr} & 0.1, 0.2, 0.3 \\
	\end{tabular}
	\vspace{-.5em}
	\caption{\small{Hyperparameters for training ViT-B, Swin-B, and Swin-L on COCO benchmark.}}
	\label{tab:impl_coco}
	\vspace{-.5em}
\end{table}

\noindent\textbf{MAE~\cite{mae} setting.} We equally divide the 12 ViT-B blocks into 4 subsets and apply convolutions to upsample or downsample the intermediate feature maps for producing different scales following~\cite{benchmarking,mae}. We train ViT-B with large-scale jitter (1024$\times$1024 resolution, scale range [0.1, 2.0])~\cite{lsj}, AdamW~\cite{adamw} with cosine learning rate decay, and drop path regularization for both 25 \& 100 epochs, as shown in Table~\ref{tab:impl_coco}. More details can be found in~\cite{benchmarking}.

\noindent\textbf{GreenMIM~\cite{greenmim} setting.} The learning rate setting is slightly different from Table~\ref{tab:impl_coco}. The peak learning rate is set to 1e-4 with a batch size of 16. The Swin-B is initialized with self-supervised pre-trained checkpoint via our \emph{FastMIM}. More details can be found in~\cite{greenmim}.

\begin{table}[!ht]
	\centering
	\small
	\setlength{\tabcolsep}{5pt}
	\begin{tabular}{y{87}|x{130}}
		config & ViT-B~\cite{vit}, Swin-B~\cite{swin} \\
		\shline
		optimizer & AdamW \cite{adamw} \\
		peak learning rate & 1e-3, 3e-4 \\
		weight decay & 0.05 \\
		optimizer momentum & $\beta_1, \beta_2{=}0.9, 0.999$ \cite{chen2020generative} \\
		layer-wise lr decay~\cite{lrdecay,beit} & 0.65, 0.9 \\
		batch size & 16 \\
		learning rate schedule & linear \\
		warmup steps & 1500 \\
		training steps & 160K  \\
		input resolution & (512, 512) \\
		drop path rate~\cite{dpr} & 0.1 \\
	\end{tabular}
	\vspace{-.5em}
	\caption{\small{Hyperparameters for training ViT-B and Swin-B on ADE20K benchmark.}}
	\label{tab:impl_ade20k}
	\vspace{-.5em}
\end{table}

\noindent\textbf{SimMIM~\cite{simmim} setting.} Table~\ref{tab:impl_coco} shows the corresponding hyperparameters for Swin-B and Swin-L following~\cite{simmim}. The window size for Swin-B is set to 7 and that for Swin-L is 14. Notably, we choose upgraded Mask R-CNN (more details in Sec.~2.2 in~\cite{benchmarking}) as basic framework and initialize the backbone with checkpoint fine-tuned on ImageNet-1K, following SimMIM~\cite{simmim}. More details can be found in~\cite{simmim}.

\subsection{Semantic Segmentation on ADE20K} We use typical UperNet~\cite{upernet} as the basic framework. We follow two previous settings to evaluate our \emph{FastMIM}.

\begin{table*}[t]
	\centering
	\tablestyle{9pt}{1.0}
	\begin{tabular}{lccccccc}
		\toprule
		\textbf{Framework} & \textbf{Model} & \textbf{\# Params} & \textbf{PT Ep.} & \textbf{Hours/Ep.} & \textbf{PT Hours} & \textbf{FT Ep.} & \textbf{Top-1 (\%)} \\
		\midrule
		Supervised training~\cite{mae}& ViT-B & 86M & 0 & 1.6\rlap{$^{\dagger,\ddagger}$} & 490\rlap{$^{\dagger,\ddagger}$} & 300 & 82.3 \\
		\baseline{FastMIM (ours)} & \baseline{ViT-B} & \baseline{86M} & \baseline{800} & \baseline{0.8} & \baseline{608} & \baseline{100} & \baseline{83.8}\rlap{~\scriptsize(+1.5)} \\
		\midrule
		Supervised training~\cite{swin} & Swin-B & 88M & 0 & 2.5\rlap{$^{\dagger,\ddagger}$} & 744\rlap{$^{\dagger,\ddagger}$} & 300 & 83.5 \\
		\baseline{FastMIM (ours)} & \baseline{Swin-B} & \baseline{88M} & \baseline{400} & \baseline{0.8} & \baseline{336} & \baseline{100} & \baseline{84.1}\rlap{~\scriptsize(+0.6)} \\
		\midrule
		Supervised training~\cite{simmim} & Swin-L & 197M & 300 & 2.0\rlap{$^\ddagger$} & 1139\rlap{$^\ddagger$} & 100 & 83.5 \\
		\baseline{FastMIM (ours)} & \baseline{Swin-L} & \baseline{197M} & \baseline{800} & \baseline{1.4} & \baseline{1088} & \baseline{100} & \baseline{85.4}\rlap{~\scriptsize(+0.9)} \\
		\midrule
		Supervised training~\cite{twins} & Twins-L & 99M & 0 & 2.8\rlap{$^{\dagger,\ddagger}$} & 832\rlap{$^{\dagger,\ddagger}$} & 300 & 83.7 \\
		\baseline{FastMIM (ours)} & \baseline{Twins-L} & \baseline{99M} & \baseline{800} & \baseline{0.9} & \baseline{716} & \baseline{100} & \baseline{84.0}\rlap{~\scriptsize(+0.3)} \\
		\midrule
		Supervised training~\cite{pvt} & PVTv1-L & 61M & 0 & 2.1\rlap{$^{\dagger,\ddagger}$} & 624\rlap{$^{\dagger,\ddagger}$} & 300 & 81.7  \\
		\baseline{FastMIM (ours)} & \baseline{PVTv1-L} & \baseline{61M} & \baseline{800} & \baseline{0.7} & \baseline{592} & \baseline{100} & \baseline{82.9}\rlap{~\scriptsize(+1.2)} \\
		\midrule
		Supervised training~\cite{pvtv2} & PVTv2-B2 & 25M & 0 & 1.7\rlap{$^{\dagger,\ddagger}$} & 504\rlap{$^{\dagger,\ddagger}$} & 300 & 82.0 \\
		\baseline{FastMIM (ours)} & \baseline{PVTv2-B2} & \baseline{25M} & \baseline{800} & \baseline{0.6} & \baseline{448} & \baseline{200} & \baseline{82.6}\rlap{~\scriptsize(+0.6)} \\
		\midrule
		Supervised training~\cite{cmt} & CMT-S & 25M & 0 & 2.8\rlap{$^{\dagger,\ddagger}$} & 840\rlap{$^{\dagger,\ddagger}$} & 300 & 83.5\\
		\baseline{FastMIM (ours)} & \baseline{CMT-S} & \baseline{25M} & \baseline{800} & \baseline{1.0} & \baseline{768} & \baseline{200} & \baseline{83.9}\rlap{~\scriptsize(+0.4)} \\
		\bottomrule
	\end{tabular}
	\vspace{-0.2cm}
	\caption{\small{Comparison with supervised training method on more vision backbones. ``PT Ep." refers to the number of pre-training epoch. ``Hours/Ep." refers to GPU hours per epoch. ``PT Hours" refers to total pre-training GPU hour. ``FT Ep." refers to fine-tuning epoch. We report top-1 accuracy on ImageNet-1K validation set. ($^\dagger$: we report the total hours in fine-tuning stage. $^\ddagger$: result tested by us.)}}
	\vspace{-0.2cm}
	\label{tab:more_backbones}
\end{table*}

\begin{figure}[t]
	\centering
	{\begin{subfigure}{\linewidth}
			\centering         
			\includegraphics[width=\linewidth]{fig/swin_b_hog_epoch.pdf}
			\\[0.5mm]
			\includegraphics[width=\linewidth]{fig/swin_b_pixel_epoch.pdf}
			\vspace{-0.4cm}
			\caption{\small{ImageNet-1K top-1 accuracy of Swin-B~\cite{swin}.}}
			\label{supp_fig:abla_on_swin}
		\end{subfigure}
	}\\[2mm]
	{\begin{subfigure}{\linewidth}
			\centering         
			\includegraphics[width=\linewidth]{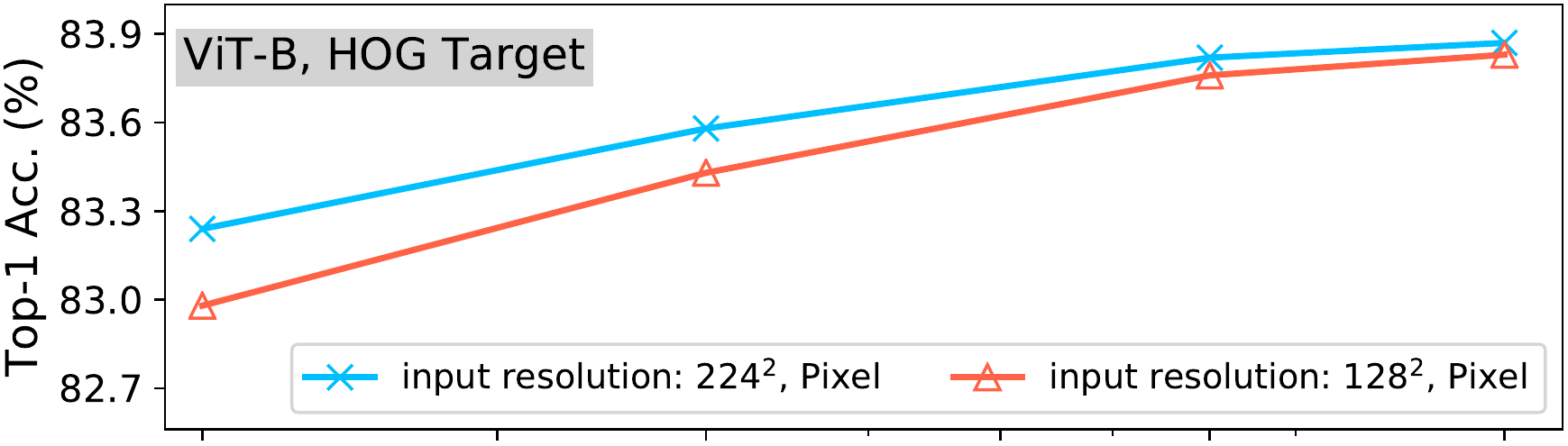}
			\\[0.5mm]
			\includegraphics[width=\linewidth]{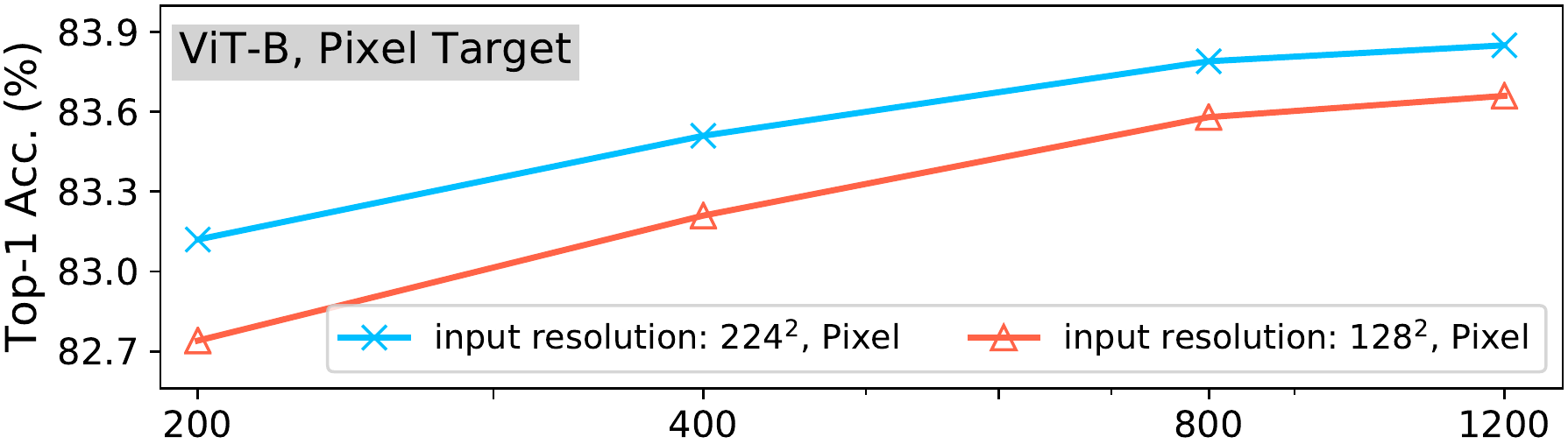}
			\vspace{-0.4cm}
			\caption{\small{ImageNet-1K top-1 accuracy of ViT-B~\cite{swin}.}}
			\label{supp_fig:abla_on_vit}
		\end{subfigure}
	}
	\vspace{-0.2cm}
	\caption{\small{Ablation on training epoch/resolution/target. A longer epoch gives a noticeable improvement. HOG can preserve better performance with low resolution input compared with pixel target.}}
	\label{supp_fig:abla_on_both}
	\vspace{-0.2cm}
\end{figure}

\noindent\textbf{MAE~\cite{mae} setting.} We follow the semantic segmentation code of MAE~\cite{mae} and BEiT~\cite{beit}. We fine-tune end-to-end for 100 epochs with a batch size of 16. We turn on relative position bias only during transfer learning, initialized as zero. We fine-tune end-to-end for 160K iterations using AdamW~\cite{adamw} optimizer with the peak learning rate of 3e-4, weight decay of 0.05. The ViT-B model is trained with input resolution of 512$\times$512, as shown in Table~\ref{tab:impl_ade20k}.

\noindent\textbf{SimMIM~\cite{simmim} setting.} We follow the setting of SimMIM~\cite{simmim}: a weight decay of 0.05, a batch size of 32, a layer-wise learning rate decay rate of 0.9, and a peak learning rate of 3e-4. The Swin-B model is trained with input resolution of 512$\times$512, as shown in Table~\ref{tab:impl_ade20k}. We initialized the backbone with checkpoint after supervised fine-tuning on ImageNet-1K. In inference, a multi-scale test using resolutions that are [0.75, 0.875, 1.0, 1.125, 1.25]$\times$ of 512$\times$2048 is employed.

\begin{table}[t]
	\centering
	\small
	\setlength{\tabcolsep}{8pt}
	\begin{tabular}{cccccc}
		ep.$\backslash$inp. & 96$^2$ & 128$^2$ & 160$^2$ & 192$^2$ & 224$^2$ \\ 
		\shline
		400 & 82.47 & 82.89 & 83.02 & 83.08 & 83.15 \\
		800 & 82.74 & 83.06 & 83.16 & 83.24 & 83.34 \\
	\end{tabular}
	\vspace{-0.2cm}
	\caption{\small{Ablation on the input resolutions. Setting: MAE~\cite{mae}, ViT-B, raw pixel as prediction target. Top-1 Acc. is reported.}}
	\vspace{-0.2cm}
	\label{supp_tab:input_resolution_vit}
\end{table}

\section{\emph{FastMIM} for Other Vision Backbones}
\label{appendix_sec:exp_other_bb}
Our proposed \emph{FastMIM} can serve as a generic MIM framework for various vision backbones, including vanilla isotropic ViT~\cite{vit}, hierarchical Swin Transformer~\cite{swin}, Twins~\cite{twins}, PVT~\cite{pvt,pvtv2}, and CMT~\cite{cmt}. We conduct experiments based on above vision backbones and compare the top-1 accuracy with  previous supervised training results. As shown in Table~\ref{tab:more_backbones}, our \emph{FastMIM} consumes fewer pre-training hours but obtains consistently better performance on all architectures. In particular, our \emph{FastMIM} achieves 84.0/82.9/82.6/83.9\% top-1 accuracy with Twins-L/PVTv1-L/PVTv2-B2/CMT-S, which is +0.3/+1.2/+0.6/+0.4\% better than the supervised training counterparts. These results demonstrate the efficiency and effectiveness of our proposed generic \emph{FastMIM} framework.

\section{Reduce Input Resolution in MAE}
\label{appendix_sec:reduce_mae}
Table~\ref{supp_tab:input_resolution_vit} ablates how the pre-training epoch and input resolution impact the fine-tuning result of MAE framework~\cite{mae}. The final performance decreases when the input resolution is reduced. However, the performance drop resulted from decreasing input resolution of MAE from  224$^2$ to 128$^2$ is slightly larger when compared with MIM~\cite{simmim}. We conjecture one main reason is that the MAE discards up to 75\% patches during pre-training stage, and reducing the input resolution will drastically decrease the number of visible patches, together with crucial position information for encoder. Although there is an extra absolute positional embedding added to the encoder input, the ability to capture (perceive) location information of MAE is inferior to MIM which retains the whole input patches.

\begin{figure*}[!h]
	\centering
	{\begin{subfigure}{\linewidth}
			\centering
			\includegraphics[width=0.48\linewidth]{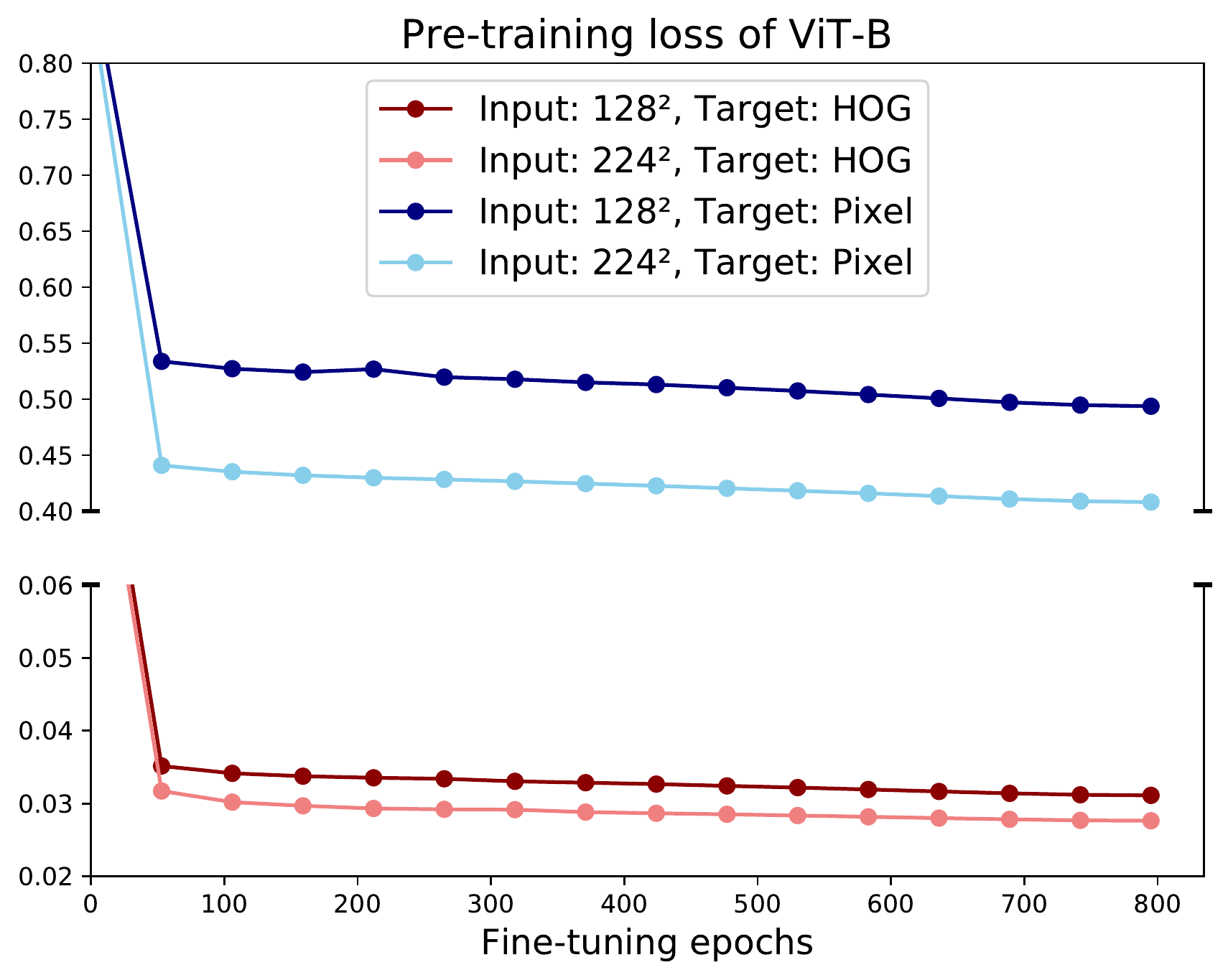}
			\hspace{2mm}
			\includegraphics[width=0.48\linewidth]{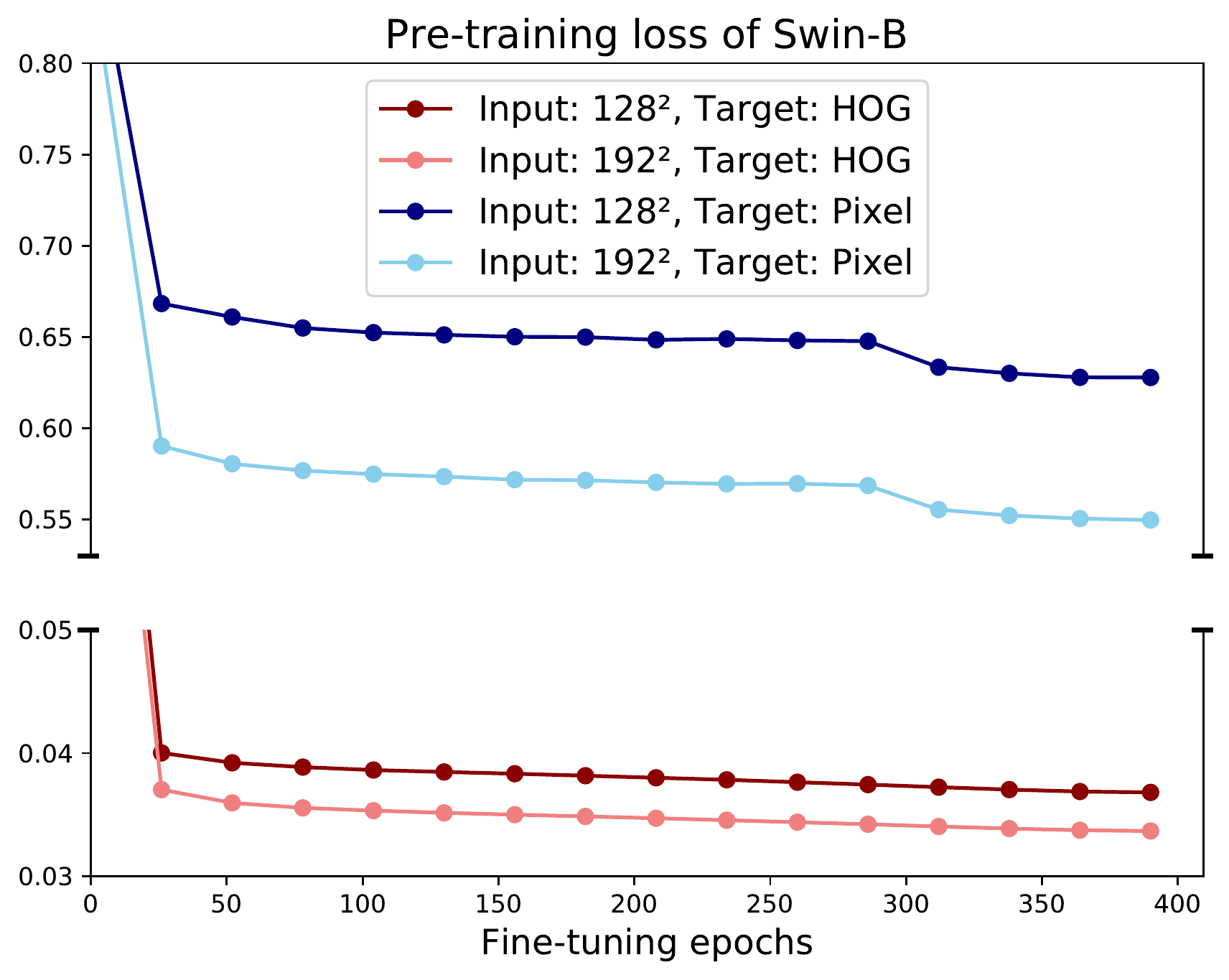} 
		\end{subfigure}
	}
	\vspace{-0.2cm}
	\caption{\small{Pre-training loss on ImageNet-1K~\cite{imagenet}. ViT-B (left) is trained
			with 800 epochs and Swin-B (right) is trained with 400 epochs.}}
	\label{supp_fig:ptloss}
	\vspace{-.1in}
\end{figure*}

\begin{figure}[t]
	\centering
	{\begin{subfigure}{\linewidth}
			\centering         
			\includegraphics[width=\linewidth]{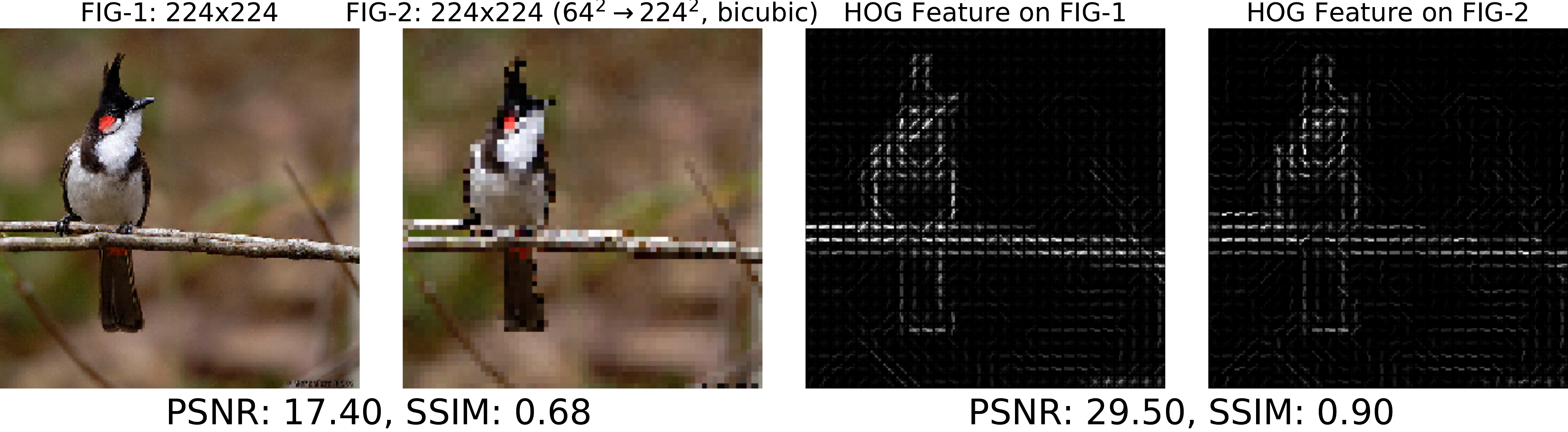}
			\\[0.5mm]
			\includegraphics[width=\linewidth]{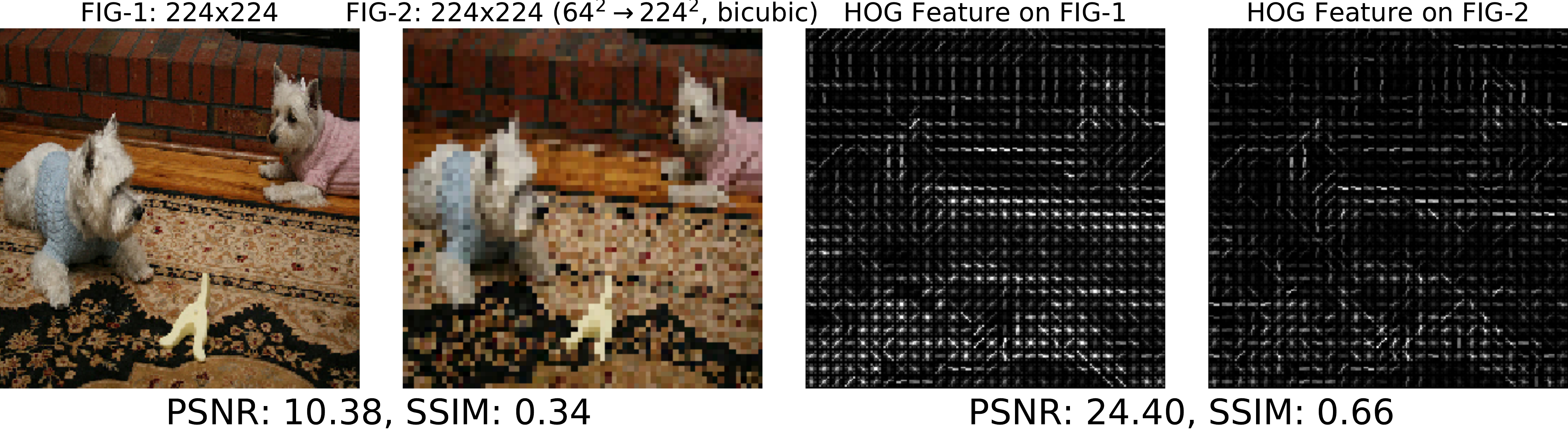}
			\\[0.5mm]
			\includegraphics[width=\linewidth]{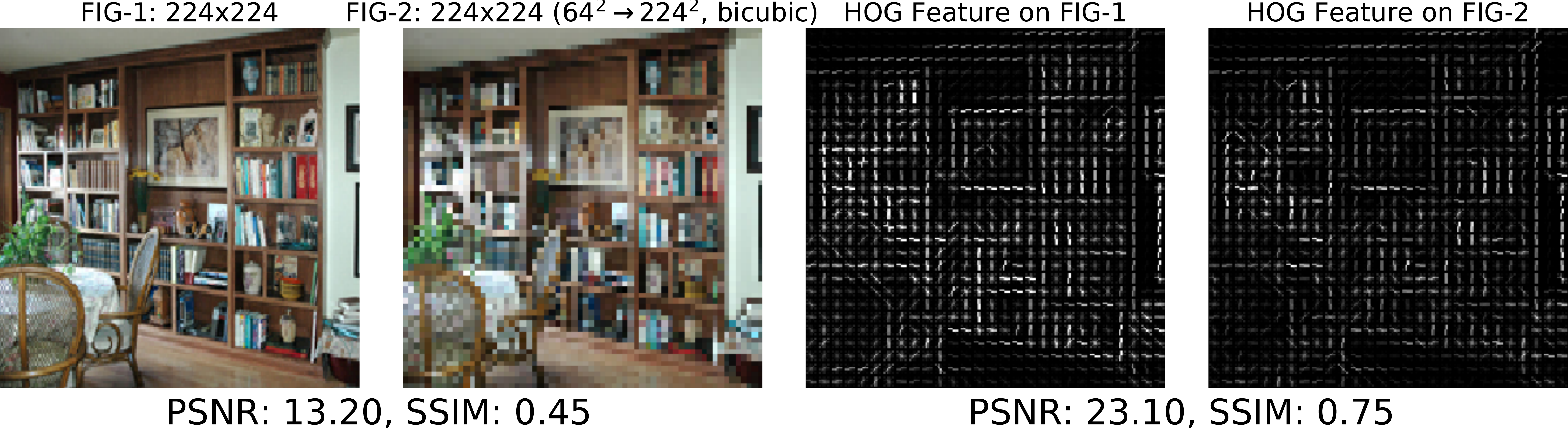}
			\\[0.5mm]
			\includegraphics[width=\linewidth]{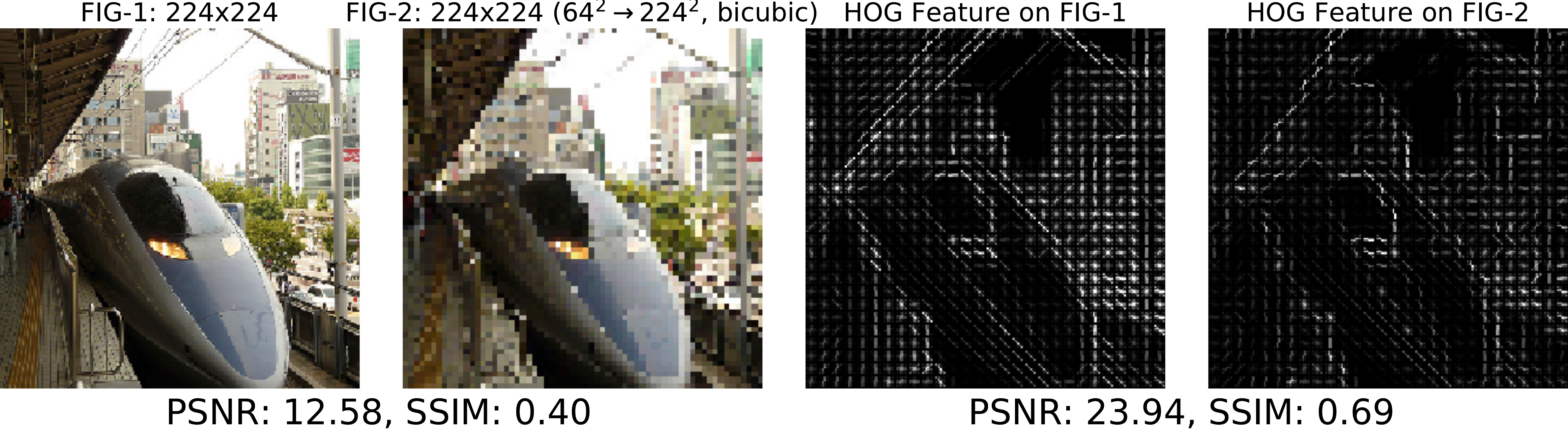}
			\\[0.5mm]
			\includegraphics[width=\linewidth]{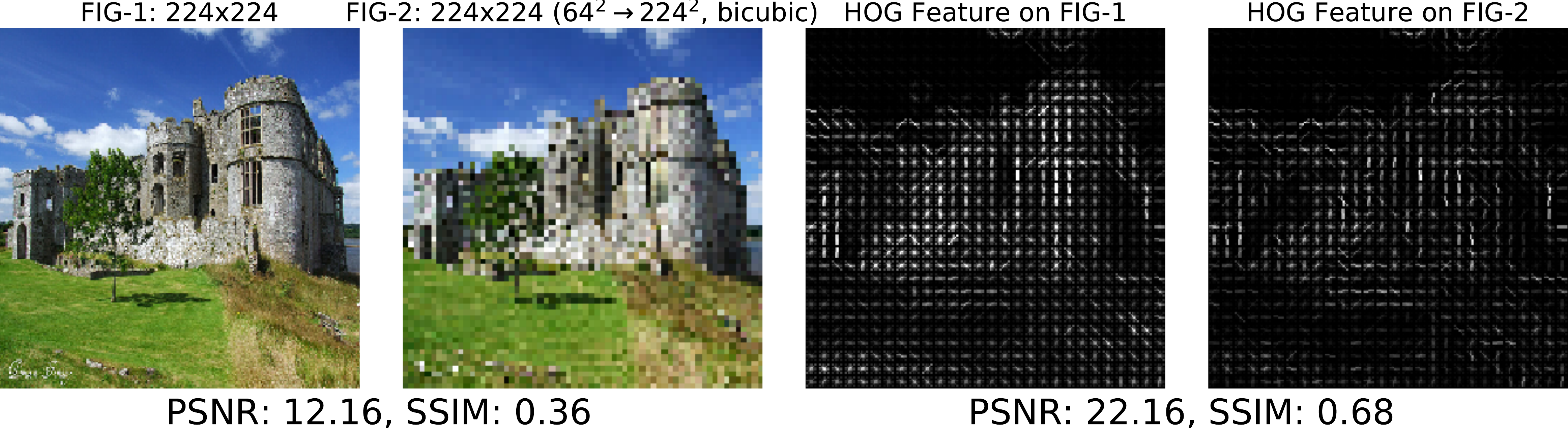}
			\vspace{-0.2cm}
		\end{subfigure}
	}
	\vspace{-0.2cm}
	\caption{\small{Visualization on pixel target and HOG target. Images are randomly chosen from ImageNet-1K~\cite{imagenet}. We choose PSNR(dB) and SSIM~\cite{psnr} to evaluate the similarity between two images (features). HOG target can preserve better texture information under low resolution input compared to pixel target.}}
	\label{supp_fig:vis_hog_pixel_in1k}
	\vspace{-.1in}
\end{figure}

\begin{figure}[t]
	\centering
	{\begin{subfigure}{\linewidth}
			\centering
			\includegraphics[width=\linewidth]{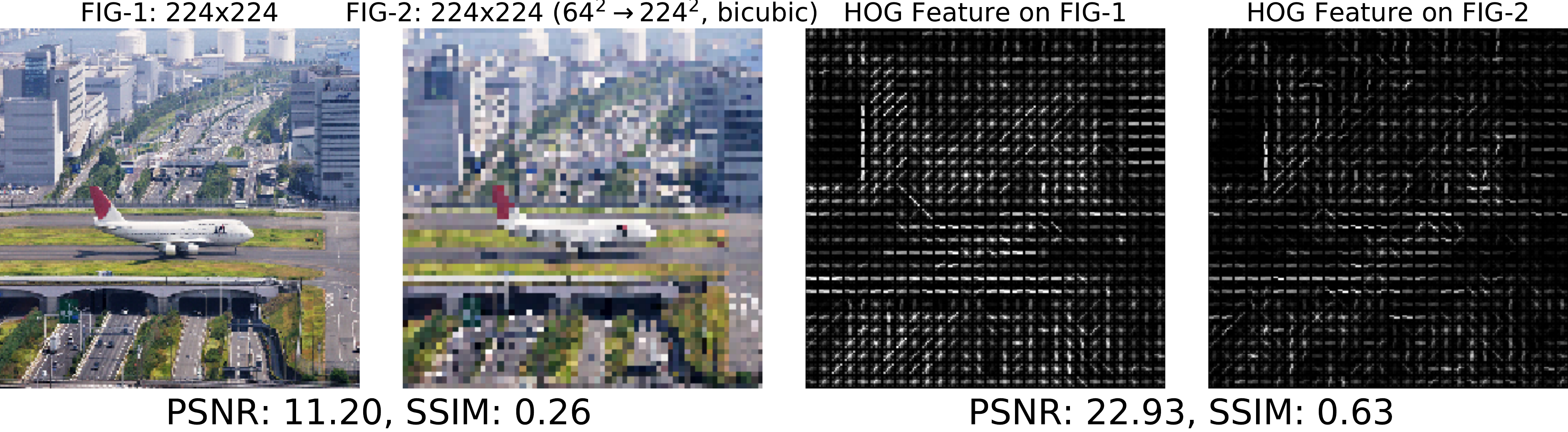}
			\\[0.5mm]
			\includegraphics[width=\linewidth]{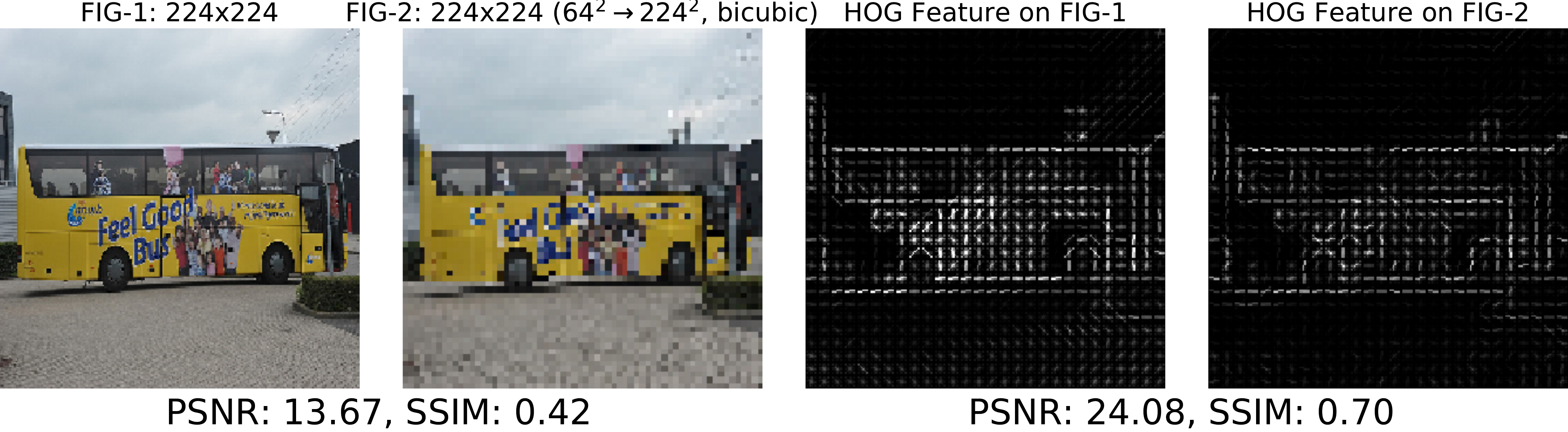}
			\\[0.5mm]
			\includegraphics[width=\linewidth]{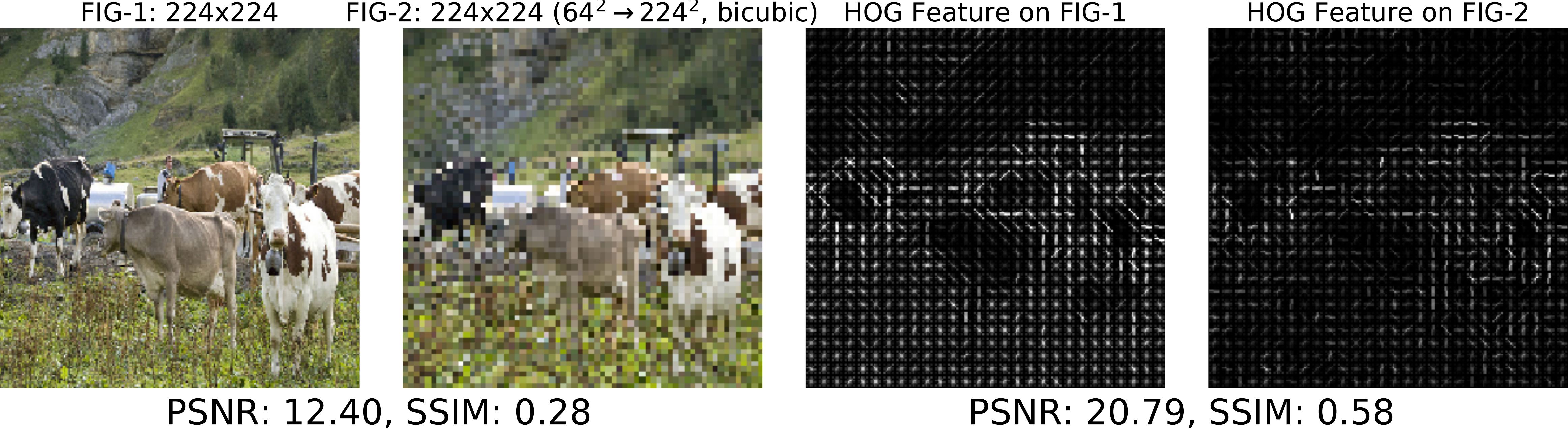}
			\\[0.5mm]
			\includegraphics[width=\linewidth]{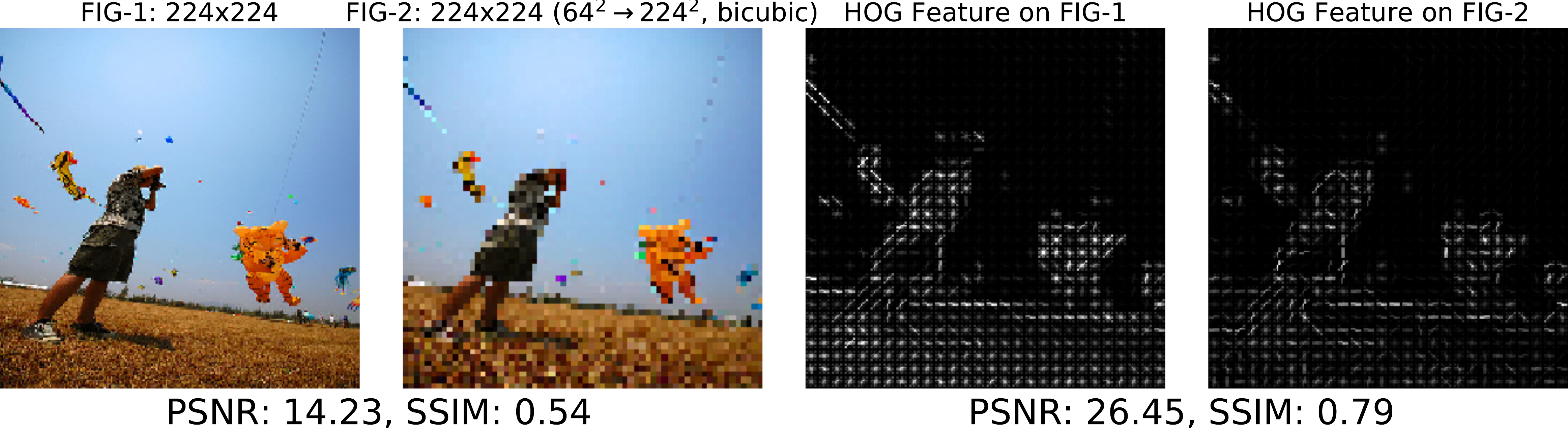}
			\\[0.5mm]
			\includegraphics[width=\linewidth]{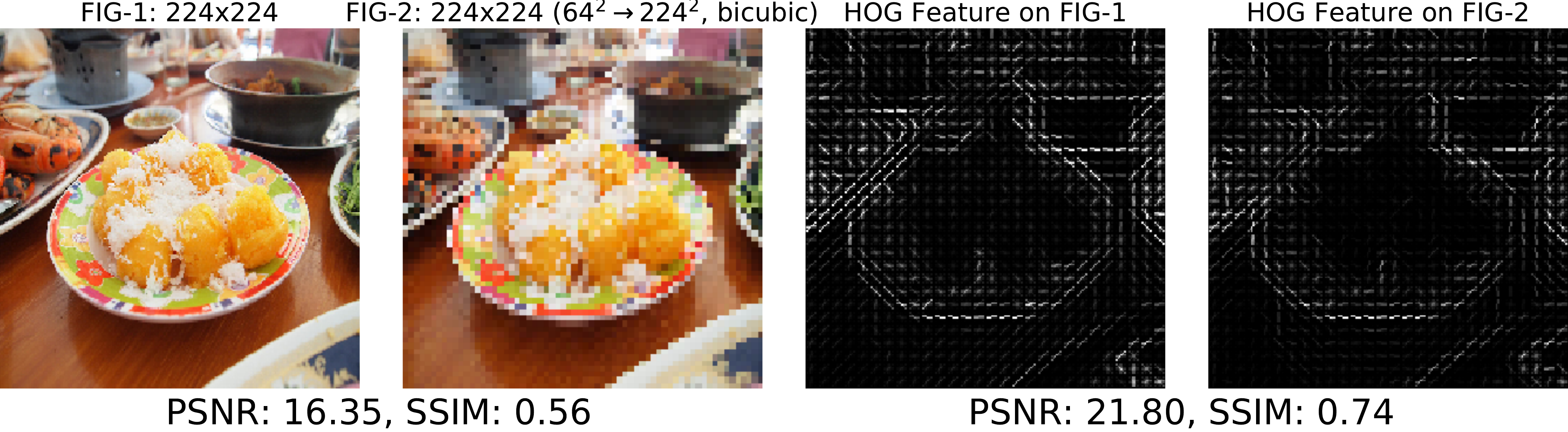}
			\vspace{-0.2cm}
		\end{subfigure}
	}
	\vspace{-0.2cm}
	\caption{\small{Visualization on pixel target and HOG target. Images are randomly chosen from COCO~\cite{coco}. We choose PSNR(dB) and SSIM~\cite{psnr} to evaluate the similarity between two images (features). HOG target can preserve better texture information under low resolution input compared to pixel target.}}
	\label{supp_fig:vis_hog_pixel_coco}
	\vspace{-.2cm}
\end{figure}

\begin{figure*}[t]
	\centering
	{\begin{subfigure}{\linewidth}
			\centering
			mask size = 32$\times$32 \\[0.75mm]
			\includegraphics[width=\linewidth]{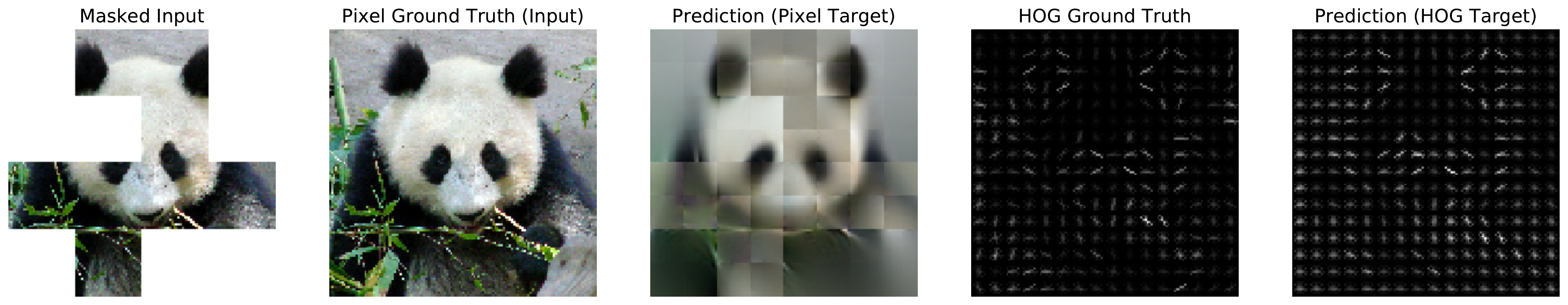}
			\\[0.15mm]
			\includegraphics[width=\linewidth]{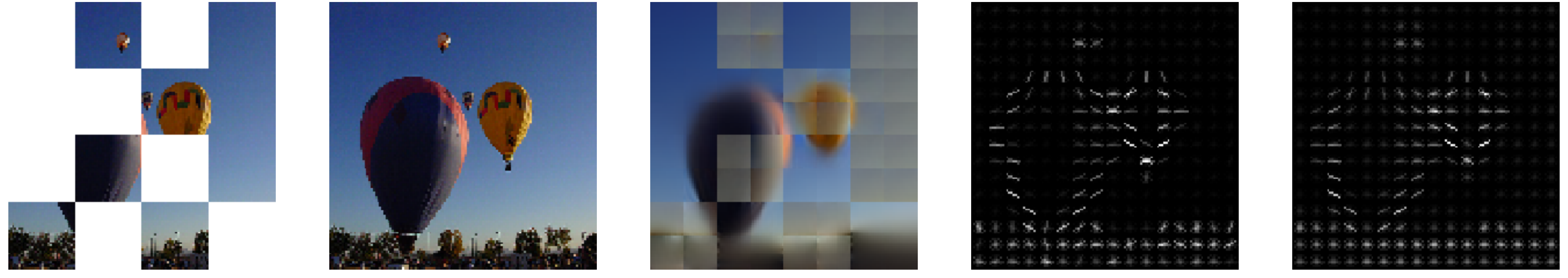}     
			\\[0.15mm]
			\includegraphics[width=\linewidth]{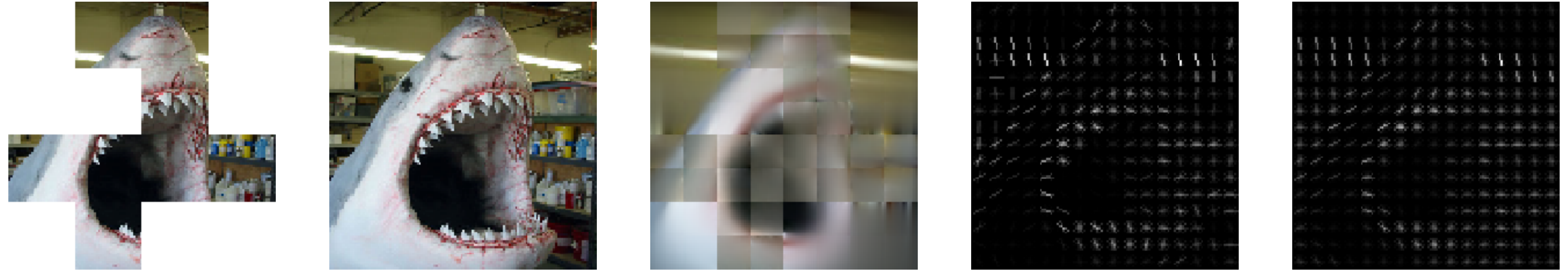}
			\\[0.15mm]
			\includegraphics[width=\linewidth]{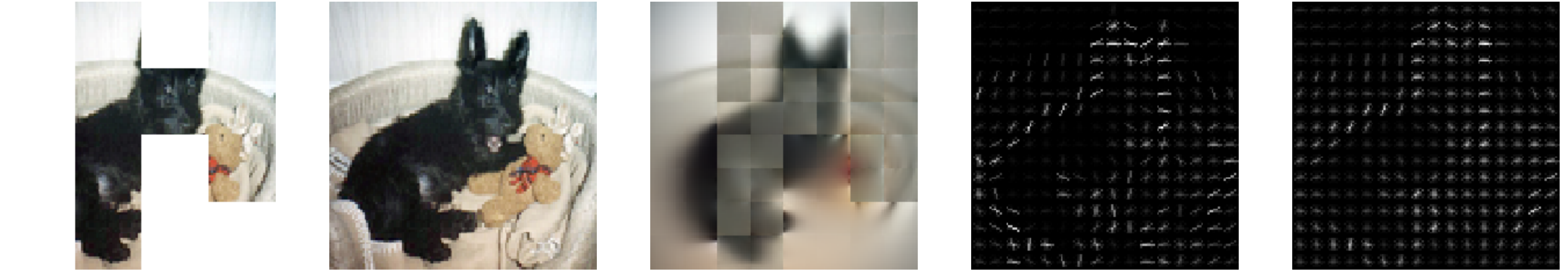}
			\\[0.15mm]
			\includegraphics[width=\linewidth]{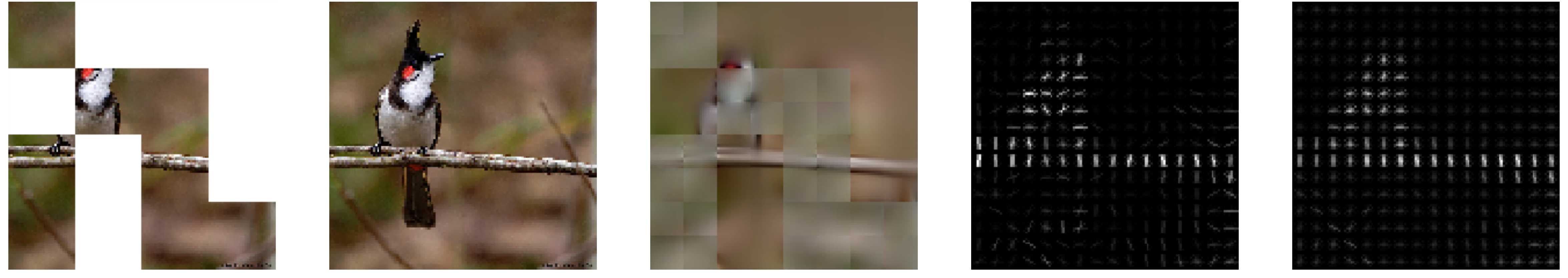}
			\\[.0mm] Failure Case \\[.75mm]
			\includegraphics[width=\linewidth]{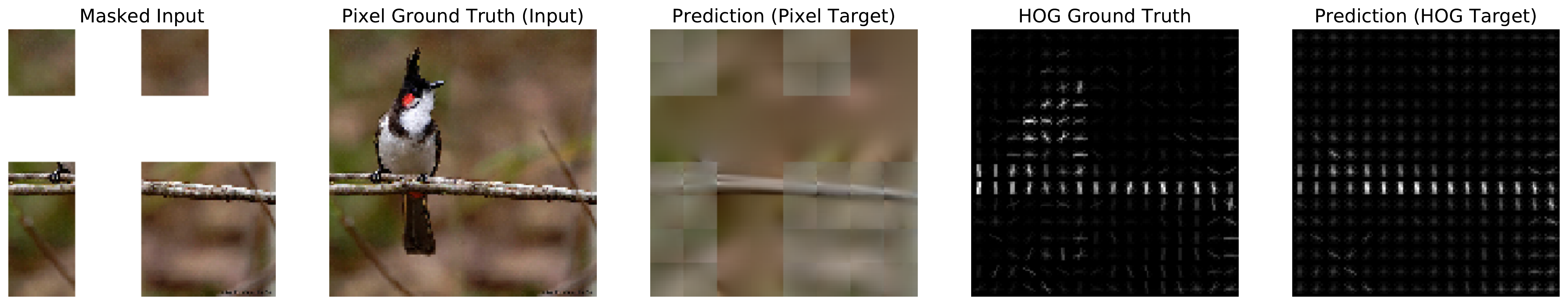}
			\vspace{-0.2cm}
		\end{subfigure}
	}
	\vspace{-0.6cm}
	\caption{\small{Pixel vs. HOG predictions (without normalization) on ImageNet-1K~\cite{imagenet} validation set. Using an MIM trained on ImageNet. For each sample, we show the masked image, original input, prediction trained by pixel target, HOG ground truth, and prediction trained by HOG target. The unmasked regions are not used for loss and thus qualitatively poor.}}
	\label{supp_fig:rec_in1k}
	\vspace{-.1in}
\end{figure*}

\begin{figure*}[t]
	\centering
	{\begin{subfigure}{\linewidth}
			\centering
			mask size = 16$\times$16 \\[0.75mm]
			\includegraphics[width=\linewidth]{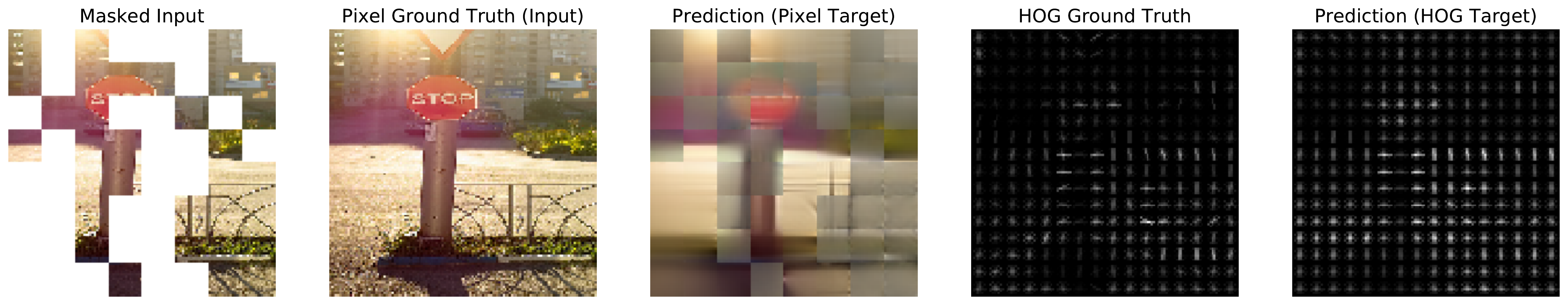}    
			\\[0.15mm]
			\includegraphics[width=\linewidth]{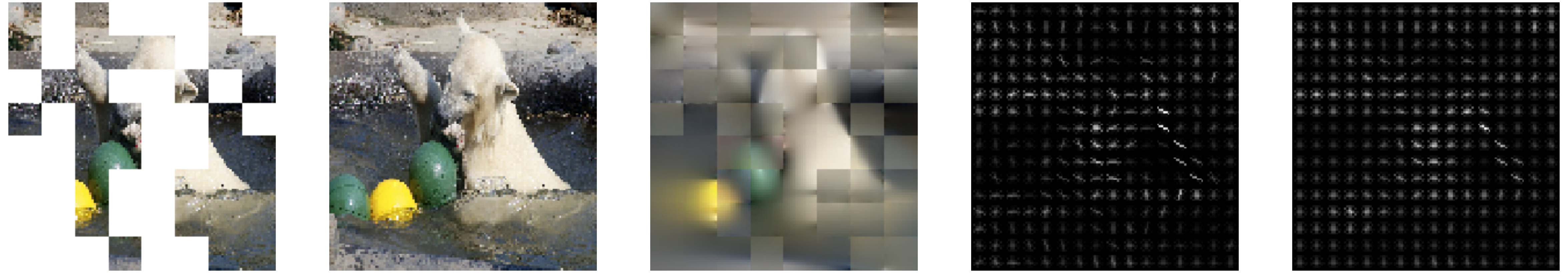}
			\\[0.15mm]
			\includegraphics[width=\linewidth]{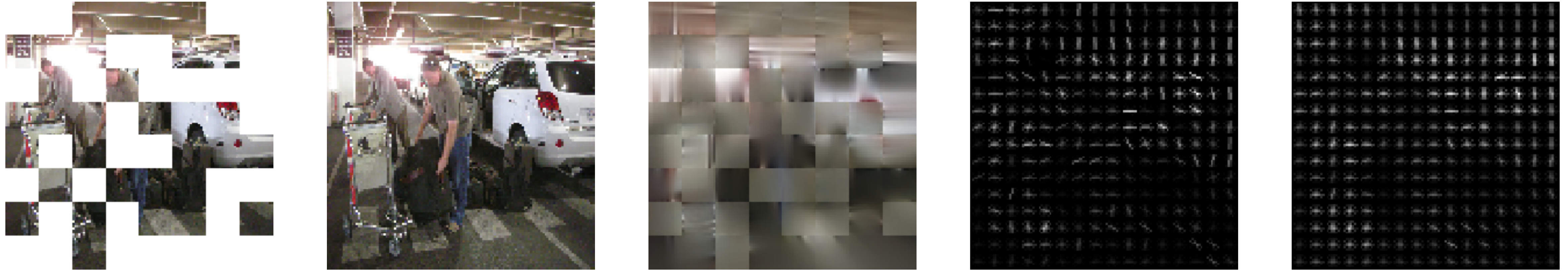} 
			\\[0.mm] mask size = 32$\times$32 \\[0.5mm]
			\includegraphics[width=\linewidth]{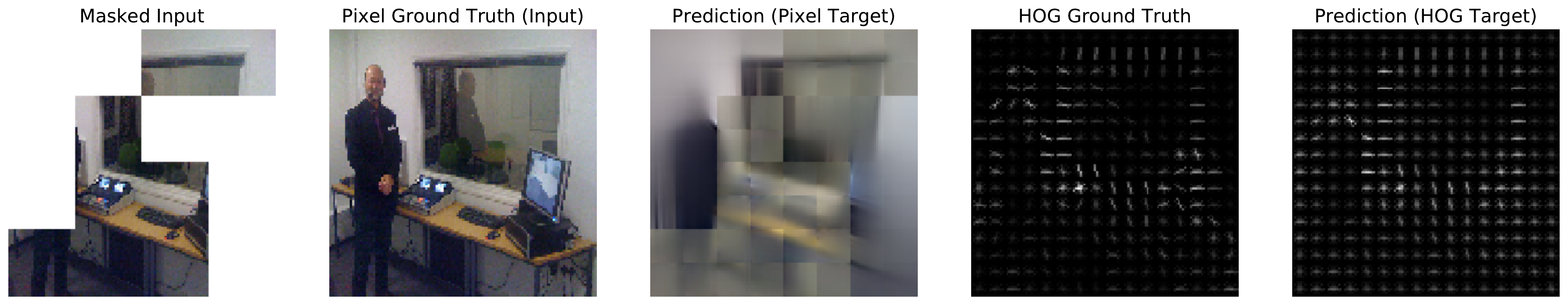}
			\\[0.15mm]
			\includegraphics[width=\linewidth]{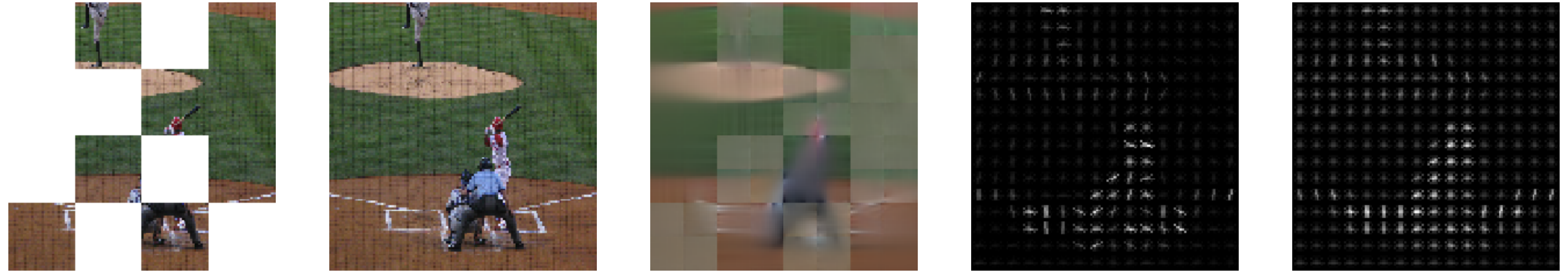}
			\\[.0mm] Failure Case \\[.5mm]
			\includegraphics[width=\linewidth]{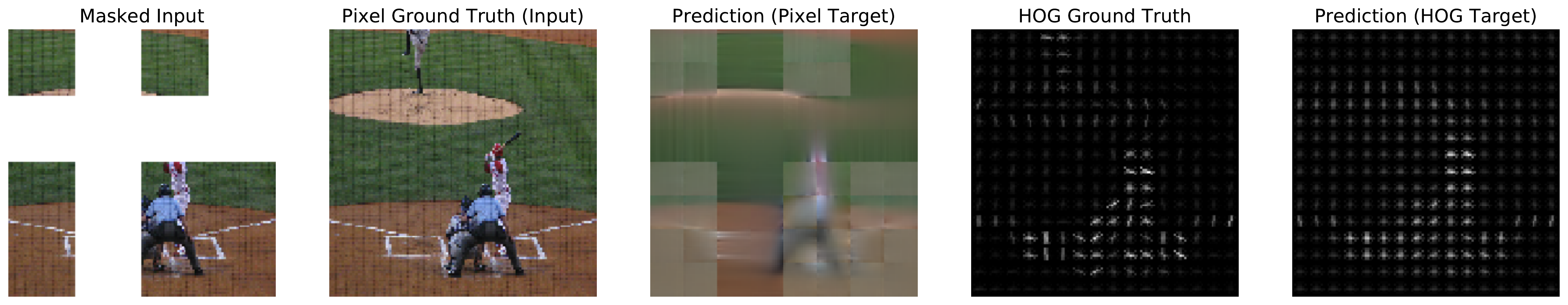}
			\vspace{-0.2cm}
		\end{subfigure}
	}
	\vspace{-0.6cm}
	\caption{\small{Pixel vs. HOG predictions (without normalization) on COCO~\cite{imagenet} validation set. Using an MIM trained on ImageNet. For each sample, we show the masked image, original input, prediction trained by pixel target, HOG ground truth, and prediction trained by HOG target. The unmasked regions are not used for loss and thus qualitatively poor.}}
	\label{supp_fig:rec_coco}
	\vspace{-.1in}
\end{figure*}

\section{More Ablations on HOG Target}
\label{appendix_sec:abla_hog}
\noindent\textbf{Input resolution \& target.}
In addition to Figure~\ref{supp_fig:abla_on_swin} shown in our main paper, we also present the influence of different training schedule lengths, input resolutions, and prediction targets based on ViT-B~\cite{vit} as shown in Figure~\ref{supp_fig:abla_on_vit}. The observation is similar to Swin-B that HOG target can better reduce the gap between different input resolutions than pixel target.

\noindent\textbf{Ground truth visualization of target.} Figure~\ref{supp_fig:ptloss} shows the pre-training losses of different input resolutions. We can find that HOG target can reduce the gap of loss values between different input resolutions. Besides, the absolute loss values of using HOG target are far smaller than those of using pixel target, demonstrating that HOG can effectively reduce the risk of ambiguity during reconstruction in MIM.

\noindent\textbf{Ground truth visualization of target.} Figure~\ref{supp_fig:vis_hog_pixel_in1k} and Figure~\ref{supp_fig:vis_hog_pixel_coco} show more visualization results of pixel and HOG target on ImageNet-1K and COCO, respectively.
Although reducing the image resolution can significantly expedite the training process, the crucial information, \eg, detailed textures and edges, will be discarded when using pixel target. However, HOG is more invariant to the resolution changes, which is suitable for our \emph{FastMIM}. 

\noindent\textbf{Pre-raining loss of HOG target.}
We qualitatively compare the reconstruction result of pixel target with HOG target as shown in Figure~\ref{supp_fig:rec_in1k} and Figure~\ref{supp_fig:rec_coco}. We can find that both pixel and HOG predictions are semantically plausible to some extent. However, pixel targets suffer from large errors caused by ambiguous problems~\cite{maskfeat}, while HOG is more robust to ambiguity. As shown in the second row in Figure~\ref{supp_fig:rec_in1k}, the model trained via pixel target predicts the balloon as dark blue, which is in fact red in the top area. This wrong prediction can result in a high loss penalty, which can also increase the difficulty of training. This is also affirmed by MaskFeat~\cite{maskfeat} and is also the main reason for MaskFeat to leverage HOG feature as the prediction target. In addition to above reason, we demonstrate that HOG is more invariant to the resolution changes when compared with pixel target. Therefore, HOG target is naturally more suitable for our \emph{FastMIM}.

\clearpage
\clearpage
{\small
\bibliographystyle{ieee_fullname}
\bibliography{egbib}
}

\end{document}